\documentclass[twoside,11pt]{article}

\usepackage{jmlr2e}
\usepackage{parskip}
\usepackage{hyperref}
\usepackage{amsmath}
\usepackage{epsfig}
\usepackage{amssymb}
\usepackage{mathrsfs}
\usepackage[normalem]{ulem}
\usepackage{amsmath}
\usepackage[]{algorithm2e}
\usepackage{cancel}
\usepackage[usenames,dvipsnames,svgnames,table]{xcolor}
\usepackage{soul}

\ShortHeadings{Safe Active Feature Selection for Sparse Learning}{}
\firstpageno{1}

\begin{document}

\title{Safe Active Feature Selection for Sparse Learning}

\author{\name Shaogang Ren \email shaogangren@email.tamu.edu \\
	\addr Department of Electrical and Computer Engineering\\
	Texas A\&M University\\
	College Station, TX 77843, USA
	\AND
	\name Jianhua Z. Huang \email jianhua@stat.tamu.edu \\
	\addr Department of Statistics\\
	Texas A\&M University\\
	College Station, TX 77843, USA
	\AND
	\name Shuai Huang \email shuaih@uw.edu \\
	\addr Department of Industrial and Systems Engineering\\
	University of Washington\\
	Seattle, WA 98195, USA
	\AND
    \name Xiaoning Qian \email xqian@ece.tamu.edu  \\
    \addr Department of Electrical and Computer Engineering\\
    Texas A\&M University\\
    College Station, TX 77843, USA}

\editor{}

\maketitle

\begin{abstract}
	We present \textbf{s}afe \textbf{a}ctive \textbf{i}ncremental \textbf{f}eature selection~(SAIF) to scale up the computation of LASSO solutions. SAIF does not require a solution from a heavier penalty parameter as in sequential screening or updating the full model for each iteration as in dynamic screening. Different from these existing screening methods, SAIF starts from a small number of features and incrementally recruits active features and updates the significantly reduced model. Hence, it is much more computationally efficient and scalable with the number of features. More critically, SAIF has the “safe” guarantee as it has the convergence guarantee to the optimal solution to the original full LASSO problem. Such an incremental procedure and theoretical convergence guarantee can be extended to fused LASSO problems. Compared with state-of-the-art screening methods as well as working set and homotopy methods, which may not always guarantee the optimal solution, SAIF can achieve superior or comparable efficiency and high scalability with the safe guarantee when facing extremely high dimensional data sets. Experiments with both synthetic and real-world data sets show that SAIF can be up to 50 times faster than dynamic screening, and hundreds of times faster than computing LASSO or fused LASSO solutions without screening.
	
\end{abstract}


\section{Introduction}
LASSO has been a powerful tool for sparse learning to analyze data sets with $p \gg n$, where $p$ is the number of covariates or features and $n$ the number of samples. LASSO screening methods provide efficient approaches to  scale up sparse learning without solving the full LASSO problems, based on either sequential or dynamic screening methods~\citep{Ghaoui2012,WangLasso,GAP,Ren2017}. However, the existing sequential screening requires the LASSO solution with a heavier regularization penalty parameter so that the range of dual variables can be estimated tightly to help effectively screen redundant features. Different from such static sequential screening methods, dynamic screening does not require the solution with the heavier penalty parameter but relies on duality gaps during optimization iterations for feature screening. To achieve high screening power, a significant number of optimization iterations have to be operated on the full-scale problems with the original high-dimensional feature set to compute the effective duality gap. Both sequential and dynamic screening require to update the original full-scale LASSO model. To further scale up the existing screening methods, we introduce an active incremental screening method---\textbf{s}afe \textbf{a}ctive \textbf{i}ncremental \textbf{f}eature selection~(SAIF)---that can overcome all these shortcomings. SAIF dynamically manipulates the feature set according to the duality gap of the sub-problems with only a small number of active features in hand. In this way we can maximally reduce the redundant computation on inactive features  that have zero coefficients in the optimal LASSO solution.  

We first review the relevant literature on screening and homotopy methods  for sparse learning, and then present the theoretical and experimental results for SAIF in the following sections.

\subsection{Sequential Screening}

Most sequential screening methods derive screening rules by leveraging the solutions to the LASSO model with a heavier regularization parameter. There are two broad categories of sequential screening methods for LASSO problems: heuristic and safe screening methods. The heuristic screening methods~\citep{Tibshirani2012, Fan2008} rely on heuristics to remove features. For example, the strong rule~\citep{Tibshirani2012} derives the screening rule based on the assumption that the absolute values of the inner products between features and the residue are non-expansive with respect to the parameter values. It is obvious that this assumption does not always hold. Such heuristic screening rules are not \textbf{safe}, meaning that they cannot guarantee that the removed features will have corresponding zero coefficients in the optimal LASSO solution to the original full-scale problem. Sequential screening methods, such as the ones proposed by~\cite{Ghaoui2012,WangLasso,WangLogistic,Ren2017}, do not take the unsafe assumptions that the heuristic screening methods use,  but try to develop safe feature screening rules based on the structure of the problem. Most of these screening methods are inspired by the seminal work by~\citet{Ghaoui2012} and derive screening rules with the help of the LASSO solution with a heavier regularization parameter. 

Assume $n$ data samples with $p$ features form an $n\times p$ design matrix $X$, with an $n \times 1$ label vector,  $\mathbf{y}$. Let the general loss function $f$ in~\eqref{eq:plasso} be $\alpha$-smooth, and $\gamma$-convex with respect to (w.r.t.) the $L_2$ norm~($\alpha>0$, $\gamma >0$), and $f^*$ is the conjugate of $f$. According to Theorem 6 in~\cite{Kakade2009OnTD}, $f^*$ is $\frac{1}{\alpha}$-convex  $\frac{1}{\gamma}$-smooth w.r.t. the $L_2$ norm.  The primal and dual forms of the general LASSO problem can be written as follows:
\begin{align}
P: \  \   &  \min_{\mathbf{\beta}}\sum_{j=1}^n f(x_{j\bullet}\beta, y_j) + \lambda || \mathbf{\beta}||_1 \label{eq:plasso}, \\
D: \  \  & \sup_{\theta} -\sum_{j=1}^n f^*(-\lambda \theta_j, y_j) \label{eq:dlasso} \\
& s.t. \quad |x_{i}^T\mathbf{\theta}| \leq 1, \quad \forall i \in \mathcal{F}, 
\end{align}
where $x_i$ is the $i$th column of $X$, i.e, the $i$th feature vector; the row vector $x_{j\bullet}$ denotes the $j$th sample; $\mathcal{F}$ denotes the index space of the features in the original LASSO problem; $\theta$ is the dual variable.  The derivation of the dual problem~\eqref{eq:dlasso} can be referred to~\cite{GAP,WangLasso}. The optimal primal and dual variable relationship is $f'(\mathbf{x}_{j\bullet}\mathbf{\beta}^*)= -\lambda \theta_j^*$, where $f'$ is the first-order derivative of $f$. With KKT (Karush-Kuhn-Tucker) conditions~\citep{Tibshirani2011,WangLasso}, we have
\begin{align}
x_i^T\theta^*   
\begin{cases}
=\mbox{sign}([\beta^*]_i)     & \mbox{if} \ [\beta^*]_i \neq 0\\
\in [-1, 1]  & \mbox{if} \  [\beta^*]_i = 0 .\label{eq:xth}
\end{cases}
\end{align}
According to~\eqref{eq:xth}, if $
|x_i^T\theta^*|<1$, we can have  $[\beta^*]_i = 0$, and $i \  \text{is an inactive feature}.$ Most existing screening methods aim to estimate a convex or ball region $B(\theta, r) = \{\theta^*\mid ||\theta^* - \theta||_2 \leq r \}$ as the range of $\theta^*$.   If $\theta^* \in B(\theta, r)$, we can write $\theta^* = \theta + \rho$ with $||\rho||_2 \leq r$. It follows that $x_i^T\theta^* = x_i^T \theta +  x_i^T\rho $, and we have
\[x_i^T \theta - ||x_i||_2 r \leq x_i^T\theta^* \leq x_i^T \theta + ||x_i||_2 r.
\] Thus
\begin{align}
 |x_i^T \theta| + ||x_i||_2 r <1  \implies i \  \text{is an inactive feature}.\label{eq:gapsrc}
\end{align}

Safe sequential screening methods estimate the ball region of the dual variables $\theta^*$ at $\lambda$ as $B(\theta^*(\lambda'), r)$. Here $\lambda' > \lambda$, and $\theta^*(\lambda')$ can be computed based on the primal-dual relation for the solution to the LASSO problem with $\lambda'$ as the regularization penalty parameter. For example, DPP (Dual  Polytope Projection)~\citep{WangLasso} takes the dual problem~\eqref{eq:dlasso} as a projection problem and relies on the estimation of $B(\theta^*(\lambda'), r)$ for the range of $\theta^*$ based on the non-expansiveness  properties. Typically, sequential screening requires to solve a sequence of LASSO problems corresponding to a sequence of descending $\lambda$'s to tighten the range estimates of $\theta^*$ to achieve the high screening power. Such a sequential procedure is suitable and efficient when solving a sequence of sparse learning problems with different regularization parameters, for example, for parameter selection by cross validation. However, sequential screening methods are not absolutely safe since there is always computational error to the solution of LASSO with a heavier penalty parameter, which is pointed out by the authors of dynamic screening methods~\citep{GAP} reviewed in the next subsection.

\subsection{Dynamic Screening}
Different from sequential screening methods that require the solution with a heavier penalty parameter, dynamic screening~\citep{GAP, Fercoq2015, Bonnefoy2015} can scale up LASSO solutions by dual variable range estimation based on the duality gap during the algorithm iterations. The ball region for $\theta^*$ is estimated based on the duality gap as a function  of the primal and dual objective function values at iterative updates~\citep{GAP, Fercoq2015}: 
\begin{equation}
\forall \theta \in \Omega_{\mathcal{F}}, \beta \in R^{p\times 1}, \  \   B\Big(\theta, \frac{2}{\lambda^2} [    P(\beta) - D(\theta)]\Big) = \Big\{\theta^* \mid ||\theta^* - \theta ||_2^2 \leq \frac{2}{\lambda^2} \big[    P(\beta) - D(\theta) \big] \Big\}.   \label{eq:dualball}
\end{equation}

Here $ \Omega_{\mathcal{F}}= \{\theta \mid |x_{i}^T\mathbf{\theta}| \leq 1, \forall i \in \mathcal{F} \}$ is the dual feasible space corresponding to the feature set $\mathcal{F}$; $\beta$ is the current estimation of primal variables; and $\theta$ is the projected feasible dual variables of $\beta$. $P(\beta)$ and $ D(\theta)$ are the primal and dual objective function values at $\beta$ and $\theta$, respectively.  The tightness of the results depends on the duality gap $[P(\beta) - D(\theta)]$, determined by the quality of iterative updates for $\beta$ and $\theta$. Dynamic screening algorithms in~\citet{GAP, Fercoq2015} iteratively update $\beta$ and $\theta$ for the original LASSO problem with the whole feature set $X$ to check the duality gap and apply screening rules to remove inactive features. Without the solution information from a heavier parameter, dynamic screening has to iterate the operations in optimization, such as sub-gradient computation, on the original whole feature set many times to gain a small duality gap. The computation cost of these operations dilute the screening benefits as the iterations have to be repeated many times to arrive at a duality gap that is small enough to achieve desired screening power.

\subsection{Homotopy Methods}
Homotopy methods have been applied for LASSO to compute the solution path when $\lambda$ varies~\citep{Efron2004,Osborne2000,Malioutov2005,Garrigues2008,Friedman2010,Zhao2017}. This type of methods rely on  a sequence of decreasing $\lambda$ values and ``warm start" (starting the active set with the solution from the previous $\lambda$) to achieve computational efficiency. Usually these methods have multiple iteration loops to incorporate the strong rule screening, active set,  and path-wise coordinate descent. The inner loop performs coordinate descent and active set management. The outer-loop goes through a sequence of decreasing $\lambda$ values and initializes the active set at each $\lambda$ with the strong rule and warm start. Since they do not utilize safe convergence stopping criteria for the active set, they may miss some of the active features in the optimal solutions to the original LASSO formulation with the corresponding $\lambda$ values. Furthermore, this type of methods do not employ any screening rule for the inner-loop sub-problems, and it may limit their scalability. 

 Besides screening and homotopy methods, working set methods~\citep{BLITZ} maintain a working set according to some violation rules and solve a sub-problem regarding the working set at each step. By estimating an extreme feasible point based on the current solution, this method constructs the working set for the next step by the constraints that are closest to the feasible point. This kind of methods also start from solving the original full-scale problem as the existing LASSO screening methods. However,  when $p\gg n$, the basic assumption of sparse learning is that most of the given features are irrelevant and should be inactive for the optimal solutions. These existing  algorithms may not be efficient due to redundant time-consuming operations on inactive features. 
 
\begin{figure}
	\vspace{-0.1in}
	\centering	\includegraphics[width=2.7in, clip=true]{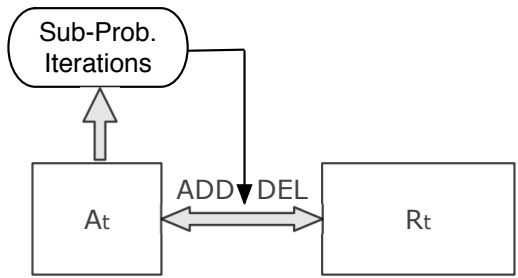}
	\caption{SAIF feature selection. $\mbox{A}_t$ stands for the Active set, while $\mbox{R}_t$ stands for the Remaining set at step $t$.}
	\label{fig:arset}
\end{figure}

\subsection{Our Contributions}

In this paper, we propose a novel safe LASSO feature selection method to further scale up LASSO solutions by overcoming the issues in the existing methods. Instead of taking the whole feature set as the initial input, our method SAIF starts from a small set of features, which is taken as the active set.  The features that are not in the active set are put in the remaining set (Figure~\ref{fig:arset}). Time-consuming iterations such as coordinate minimization with soft-thresholding are only performed on the features in the active set. Features are actively recruited or removed from the active set according to the estimated ranges of optimal dual variables. Based on duality properties, efficient feature operation rules and safe stopping criteria have been developed to keep most inactive and redundant features out of the active set. With a small active set, CPU time and memory operations can be tremendously reduced. Complexity analysis is provided for both dynamic screening and SAIF for comparison. Theoretical results show that SAIF converges to the optimal solution to the original LASSO problem and the running time of SAIF is only proportional to the active feature size, the number of features with non-zero coefficients in the optimal LASSO solution, rather than the original input feature size. Experiments on simulated and real-world data sets verified the advantages of the proposed method. Moreover we show that the proposed method can be extended to fused LASSO with tree dependency structures among features.

\begin{algorithm}[H]
	\caption{SAIF Algorithm}\label{alg:saif}
	\KwData{Data matrix $X$, label $\mathbf{y}$, penalty $\lambda$, stopping duality gap $\epsilon$ }
	\KwResult{Coefficient Vector $\beta$ }
	Choose $\lceil c\log(\frac{md+mx}{\lambda})\log(p) \rceil$ features from $\mathcal{F}$ in the descending order of $|X^T\mathbf{f}'(\mathbf{0})|$\;
	$\delta = \frac{\lambda}{\lambda_{max}}$, 
	IsAdd = True\;
	\While{True}
	{
		Update $\beta_t$ with  K iterations of soft-thresholding operations on $\mathcal{A}_t$\;
		Compute a ball region $B(\theta_t, r_t)$ based on~\eqref{eq:ball2} or~\eqref{eq:ballB};
		
		$r_t = \delta r_t$;
		
		\If{IsAdd = False \& Duality Gap $<\epsilon$}{Stop;}
		
		DEL operation;
		
		\eIf{IsAdd = False}{Continue;}
		{
			\If{$\max_{i\in R_t} |x_i^T\theta_t| + ||x_i||_2r_t < 1$}
			{
				\eIf{$\delta < 1$}
				{ $\delta = \min(10\delta, 1)$ }
				{
					IsAdd = False; Continue;
				}
			}
			ADD operation;
		}
	}
	Put $\beta_t$ in to $\beta$, and inflate the other entries with 0.
	\vspace{0.05in}
\end{algorithm}

\section{Safe Active Incremental Feature (SAIF) Selection for LASSO}

 We derive an innovative incremental feature screening algorithm, SAIF, in which we can iteratively solve much smaller sub-problems than the original LASSO problem, i.e., iteratively update the duality gap while adding or removing features by leveraging the active ball region estimates for the optimal dual variables of these sub-problems. The schematic illustration of SAIF is given in Figure~\ref{fig:arset}. Let $\mathcal{A}_t$ and $\mathcal{R}_t$ denote the active feature index set and remaining feature index set at iteration step $t$, respectively. Instead of solving either the original full-scale LASSO primal problem or the corresponding dual problem, SAIF screening is different from the existing sequential and dynamic screening as it only needs to solve significantly reduced sub-problems  and  updates the screening rules based on the duality gap without solving these sub-problems exactly.  More importantly, SAIF has the safe guarantee that only irrelevant or redundant features in the original LASSO problem will be removed. Algorithm 1 summarizes our SAIF screening procedure, which starts with $\mathcal{A}_0$ and dynamically moves active features between $\mathcal{R}_t$ and $\mathcal{A}_t$.

\subsection{ADD and DEL Operations}
Two operations in SAIF are ADD and DEL. Starting from an initial active set $\mathcal{A}_0$, whose features can be selected by some simple heuristics, for example, based on their correlation with the output, SAIF iteratively adds features (ADD) into or removes features (DEL) from the active set. At the $t$th iteration, we derive both ADD and DEL operations to dynamically update $\mathcal{A}_t$ based on the primal sub-problem with only the current active features: 
\begin{align}
P_t: \  \  & \min_{\mathbf{\beta} \in R^{|\mathcal{A}_t|}}\sum_{j=1}^n f(\sum_{i: i\in \mathcal{A}_t}x_{ji}\beta_i, y_j) + \lambda || \mathbf{\beta}||_1.  \\
D_t: \  \  & \sup_{\theta} -\sum_{j=1}^n f^*(-\lambda \theta_j, y_j) \label{eq:dlassot} \\
& s.t. \quad |x_{i}^T\mathbf{\theta}| \leq 1, \quad \forall i \in \mathcal{A}_t, \nonumber
\end{align}
Let $\Omega_{\mathcal{A}_t}$ be the dual feasible region and $D(\theta_t)$ denote the dual objective function value of the sub-problem at the dual variable $\theta_t$ considering only the active features in $\mathcal{A}_t$ with $\theta^*_t$ being the corresponding optimal dual solution. We use $\beta_t  \in R^{|\mathcal{A}_t|}$ to represent the updated $\beta$ values after $t$ out-loop iterations in SAIF. $\mathbf{\beta}^*_t$ denotes the optimal active feature solution to the problem $P_t$. $P_t(\mathbf{\tilde{\beta}})$ is the objective value of $P_t$ with input $\mathbf{\tilde{\beta}}$, and  $\mathbf{\tilde{\beta}}$ can have a different set of features from $\mathcal{A}_t$; we inflate the missing entries in $\mathbf{\tilde{\beta}}$ with zeros and ignore the entries or features not in $\mathcal{A}_t$  in the calculation of $P_t(\mathbf{\tilde{\beta}})$. Let $S_\mathcal{A}$ represent the set of the optimal primal solutions for any feature set $\mathcal{A}$,  $\theta^*$ the optimal dual solution with the full feature set $\mathcal{F}$, and $\bar{\mathcal{A}}$ for the optimal active feature set that  $\{i: |x_i^T \theta^*|=1\}$. Let $B(\theta_t, r_t) = \{ \theta_t^*\big|\  ||\theta_t^* - \theta_t||_2 \leq r_t \}$ be an estimated ball region for $\theta_t^*$ at step $t$.  

SAIF carries out ADD and DEL operations as follows:

\noindent\textbf{DEL}:
$\,\,$ For $i \in \mathcal{A}_t$, if $|x_i^T\theta_t| + ||x_i||_2 r_t < 1$, move $i$ from $\mathcal{A}_t$ to $\mathcal{R}_t$.  

\noindent\textbf{ADD}: $\,\,$ For $i \in \mathcal{R}_t$,  
if $\forall \hat{i} \in \mathcal{R}_t, \hat{i}  \neq i$, $\big||x_i ^T \theta_t| - ||x_i||_2 r_t\big| > |x_{\hat{i} }^T \theta_t|+||x_{\hat{i} }||_2r_t$, move $i$ to $\mathcal{A}_t$.

We have the following theorem regarding ADD and DEL operations: 

\noindent\textbf{Theorem 1} {\it Assume $B(\theta_t, r_t) = \{ \theta_t^*\big|\  ||\theta_t^* - \theta_t||_2 \leq r_t \}$, an estimated ball region for $\theta_t^*$ at step $t$.

a) If we add a new feature into $\mathcal{A}_t$, then $\mathcal{A}_t \subseteq \mathcal{A}_{t+1} $, $\Omega_{\mathcal{A}_t} \supseteq \Omega_{\mathcal{A}_{t+1}}$, and $D(\theta^*_{t+1}) \leq D(\theta^*_{t})$.

b) If $\exists i \in  \mathcal{R}_t$ and $|x_i^T \theta_t^*| >1$, we add feature $i$ to $\mathcal{A}_t$ at step $t$, then $D(\theta_t^*) >D(\theta_{t+1}^*)$.

c) At step $t$, if $\max_{i \in \mathcal{R}_t} |x_i^T \theta_t^*| < 1$, then $\theta_t^* = \theta^*$, $\beta_{t}^* \in S_{\mathcal{F}}$.

d) If $x_i$ satisfies $\forall \hat{i}  \in \mathcal{R}_t, \hat{i} \neq i, \big||x_i^T \theta_t| - ||x_i||_2r_t\big| \geq   |x_{\hat{i}}^T \theta_t|  +||x_{\hat{i}}||_2r_t $, then $|x_i^T \theta_t^*| \geq |x_{\hat{i}}^T\theta_t^*|, \forall  \hat{i} \in \mathcal{R}_t, \hat{i}\neq i$. }

\noindent\textbf{Proof:} a) From the dual form~\eqref{eq:dlassot}, if we add $i$ to $\mathcal{A}_t$, there will be one more constraint for the dual problem at step $t+1$, thus $\Omega_{\mathcal{A}_{t+1}} \subseteq \Omega_{\mathcal{A}_{t}}$. As we have a smaller feasible space at $t+1$, $D(\theta^*_{t+1}) \leq D(\theta^*_{t})$.

b) As $\Omega_{\mathcal{A}_{t+1}} \subset \Omega_{\mathcal{A}_{t}} $, we have $D(\theta_{t+1}^*) \leq D(\theta_t^*) $. With $|x_i^T \theta_t^*| >1$ and $|x_i^T \theta_{t+1}^*| \leq 1$,  $\theta_{t+1}^* \neq \theta_t^*$. As $\Omega_{\mathcal{A}_t}$ is convex and closed, and $\mathbf{f}^*$ is convex and smooth, the optimal dual solution for the active set $\mathcal{A}_t$ is unique, which means $D(\theta_t^*) \neq D(\theta_{t+1}^*)$. Hence, $D(\theta_t^*) > D(\theta_{t+1}^*) $. 

c) According to a), with $\mathcal{A}_t \subseteq \mathcal{F}$, we have $\Omega_\mathcal{F} \subseteq \Omega_{\mathcal{A}_t}$, and $D(\theta^*) \leq D(\theta_t^*)$. As $\forall i \in \{ \mathcal{R}_t = \mathcal{F} - \mathcal{A}_t \}, |x_i^T \theta_t^*| < 1$, $\theta_t^* \in \Omega_{\mathcal{F}}$. With $\theta^* = \sup_{\theta \in \Omega_\mathcal{F}}D(\theta)$, we get $D(\theta^*) \geq D(\theta_t^*)$. As we already know $D(\theta^*) \leq D(\theta_t^*)$, we then have $D(\theta^*) = D(\theta_t^*)$. Since the dual problem is convex and smooth, and the feasible set is closed and convex, $\theta_t^* = \theta^*$. Hence, $\beta_{t}^* \in S_{\mathcal{F}}$ as the primal solution may not be unique.

d) For ADD operations, we choose a feature in $\mathcal{R}_t$ that is the most correlated to the residual dual variables, with $\max_{i \in \mathcal{R}_t} |\mathbf{x}_i ^T \theta_t^*|$. With feature $i\in  \mathcal{R}_t$ and  $\theta^*_t \in B(\theta_t, r_t)$, we have  $\big| |x_i^T \theta_t| - ||x_i||_2 r_t \big| \leq |x_i^T\theta_t^*| \leq |x_i^T \theta_t| + ||x_i||_2 r_t$ by the Pythagorean theorem. Thus $\forall \hat{i} \in \mathcal{R}_t, \hat{i} \neq i, \big||x_i^T \theta| - ||x_i||_2r_t\big| \geq   |x_{\hat{i}} ^T \theta|  +||x_{\hat{i}}||_2r_t $, and $|x_i^T \theta_t^*| \geq |x_{\hat{i}}^T\theta_t^*|, \forall  \hat{i} \in \mathcal{R}_t, \hat{i}\neq i$.

\textbf{Remark 1} {\it Theorem 1-c) provides us with the stopping criterion for ADD operations in our SAIF algorithm. We can apply ADD and DEL operations in iterations to  minimize $\max_{i \in \mathcal{R}_t} |x_i^T \theta_t^*|$ until $\max_{i \in \mathcal{R}_t} |x_i^T \theta_t^*| < 1$. Hence, with $B(\theta_t, r_t) = \{ \theta_t^*\big|\  ||\theta_t^* - \theta_t||_2 \leq r_t \}$, if we have $\max_{i \in \mathcal{R}_t}  |x_i^T \theta_t| + ||x_i||_2 r_t <1$, we can stop ADD operations. }

\textbf{Remark 2} {\it \ Moreover, if \ $ \forall i \in \mathcal{R}_t$, $ |x_{i}^T\theta_t^*|< 1$, from Theorem 1-c), we can see that  $\theta_t^* = \theta^*$, thus $\bar{\mathcal{A}} \subseteq \mathcal{A}_t $. So  if $\bar{\mathcal{A}} \nsubseteq \mathcal{A}_t $,  $\exists i \in \mathcal{R}_t, |x_i^T\theta_t^*| \geq 1$. This concludes that  our stopping criterion for ADD operations ensures the safe feature screening. }

The DEL operation is similar to the screening steps in dynamic screening. As we can see, at step $t$ with the DEL operation, $D(\theta^*_t) = D(\theta^*_{t-1})$. Theorem 1-a) implies $D(\theta^*_t) \leq D(\theta^*_{t-1})$. Thus the optimal dual objective value always goes down.  Theorem 1-c) and Remark 1 suggest that after the stopping of ADD operations, $\mathcal{A}_t$ already has all the active features for the original problem. The algorithm then stops once it reaches the pre-specified accuracy value of the duality gap. Such monotonicity leads to the convergence of SAIF detailed in Section 3.

\subsection{Implementation}

We first discuss how we derive a tighter ball region $B(\theta_t, r_t)$ for  the range estimate of  $\theta_t^*$, taking the advantages of the existing screening methods.

\noindent\textbf{Dual variable range estimation}: Accurately estimating the range of $\theta_t^*$, $B(\theta_t, r_t)$, for the sub-problem is critical for efficient SAIF screening with ADD and DEL operations at each iteration. With $\mathbf{f}$ as the vector form of the loss function with all of the samples, we provide the following theorem to  estimate the ball region for $\theta_t^*$ with the similar idea from sequential screening. 

\noindent\textbf{Theorem 2} {\it For the LASSO problem~\eqref{eq:plasso} with the loss function $\mathbf{f}$, if $\mathbf{f}^*$ is $\frac{1}{\alpha}$-strongly convex,  and $\theta^*_0$ and $\theta^*$ are the optimal solutions to the dual problem~\eqref{eq:dlasso} at $\lambda_0$ and $\lambda$ with $\lambda < \lambda_0$, then 
\begin{equation}
||\theta^* - \frac{\lambda_0}{\lambda}\theta^*_0||_2^2 \leq \frac{2\alpha}{ \lambda^2} \bigg[ \mathbf{f}^*(-\frac{\lambda^2}{\lambda_0}\theta^*_0) - \mathbf{f}^*(-\lambda_0\theta^*_0) + (\lambda -\lambda_0)\langle  \mathbf{f}'^{*}(-\lambda_0\theta^*_0), \theta^*_0 \rangle \bigg]. 
\end{equation}
If we have $\theta \in \Omega_F$, the bound can be further improved by 
\begin{equation}
||\theta^* - \frac{\lambda_0}{\lambda}\theta^*_0||_2^2 \leq \frac{2\alpha}{ \lambda^2} \bigg[ \mathbf{f}^*(-\lambda \bar{\theta}(\bar{\varrho})) - \mathbf{f}^*(-\lambda_0\theta^*_0) + (\lambda -\lambda_0)\langle  \mathbf{f}'^{*}(-\lambda_0\theta^*_0), \theta^*_0 \rangle \bigg],
\end{equation}
where $\bar{\theta}(\bar{\varrho}) = (1-\bar{\varrho})\theta + \bar{\varrho} \frac{\lambda}{\lambda_0}\theta_0^*$, with $\bar{\varrho} = argmin_{\varrho: 0 \leq \varrho \leq 1} \mathbf{f}^*(-\lambda\bar{\theta}(\varrho))$. }

The proof for Theorem 2 can be found in Appendix A. At step $t$ with the active set $\mathcal{A}_t$, $\lambda_{max(t)}$ is the minimum $\lambda$ that leads to $\beta^*_{t}=0$. It is easy to compute $\lambda_{max(t)} = \max_{i \in \mathcal{A}_t}|x_i^T\mathbf{f}'(0)|$, and $\theta_{0(t)}^* = -\frac{\mathbf{f}'(0)}{\lambda_{0(t)}}$. If we take  $\lambda_{0(t)} = \lambda_{max(t)}$, we can use Theorem 2 to estimate $\theta_t^*$. For linear regression, the estimation can be further improved based on the projection properties as in DPP~\citep{WangLasso}.

Theorem 2 provides a tight estimation when $\lambda_0$ is close to $\lambda$. When $\lambda$ is far away from $\lambda_0$, we can adopt the tighter dual variable range estimation with the following ball region by dynamic screening~\citep{GAP,Fercoq2015}. At step $t$, we have
\begin{equation}  \label{eq:ball2}
\forall \theta_t \in \Omega_{\mathcal{A}_t}, \beta_t \in R^{p_t \times 1}, \  \   ||\theta^*_{t} - \theta_t ||_2^2 \leq \frac{2}{\lambda^2} \bigg[    P_t(\beta_t) - D(\theta_t) \bigg].  
\end{equation}
For $\beta_t$, with the primal-dual relation,  we can project it to the dual feasible region $\Omega_{\mathcal{A}_t}$ to get a feasible dual variable $\theta_t$.

With two ball regions from Theorem 2 and the duality gap, we can derive a tighter constrained region by computing the center and radius of a ball region $B(\theta_t, r_t)$ that covers the intersection of two ball regions, $B_1(\theta_1, r_1)$ and $B_2(\theta_2, r_2)$:   
\begin{align} \label{eq:ballB}
&r_t = \frac{2A}{d}, \ \ \ \ \ \theta_t = (1- \frac{d_1}{d})\theta_1 + \frac{d_1}{d} \theta_2, \ \ \ \ \   d_1 = \sqrt{r_1^2 - r_t^2} \\ \notag
& d = ||\theta_1 - \theta_2||_2, \ \ \   A = \sqrt{s(s-r_1)(s-r_2)(s-d)},  \ \ \   s = \frac{r_1 + r_2 + d}{2},
\end{align} 
where $B_1$ can be derived from Theorem 2, and $B_2$ from~\eqref{eq:ball2}. The resulting
$B(\theta_t, r_t)$ gives us a tighter bounded region at step $t$ when $r_t < \min\{r1,r2\}$. When we do not have the solutions with other $\lambda$ values, we simply set the bounded region for $\theta_t^*$ based on~\eqref{eq:ball2}.

\noindent{\textbf{Improve SAIF with an estimation factor}}: The estimation of dual variables  may be inaccurate to have high enough screening power during the optimization iterations, especially at the beginning of the algorithm. We add a factor to the radius of the ball region to reduce redundant computation resulted from inaccurately recruited features. At the beginning of Algorithm~\ref{alg:saif}, $\delta$ is a value smaller than 1.  $\delta$  will be increased to 1  during the SAIF iterations to ensure the safe  guarantee of SAIF algorithm.

\noindent{\textbf{ADD operation implementation details}}: The number of added features in each ADD operation can vary to reduce redundant iterations. Generally, the relationship between the screening power and this number depends on the regularization parameter $\lambda$ and how well feature vectors $x_i, \ i\in \mathcal{F}$, correlate with the outcome label $\mathbf{y}$. In this paper, we empirically set the number to be
$h = \lceil c\log(\frac{md+mx}{\lambda})\log(p) \rceil$. Here $mx$ and $md$ are the maximum and median of $|X^T\mathbf{f}'(\mathbf{0})|$ ($|X^T\mathbf{y}|$ with linear regression). Many iterations may need to be operated to reach the dual space point that can distinguish $h$ features, and this may reduce the efficiency of the algorithm. We can decrease the redundancy by relaxing the strict condition in Theorem 1-d). Let $V_i$ represent the set of features that violate the condition in Theorem 1-d) regarding feature $i$, i.e., $V_i = \{ \hat{i}\big| \hat{i} \in \mathcal{R}_t, \hat{i} \neq i, \big||x_{i}^T \theta_t| - ||x_{i}||_2 r_t\big| \leq |x_{\hat{i}}^T \theta_t|+||x_{\hat{i}}||_2r_t\}$. For a feature $i \in R_t$, if $|V_i|< \tilde{h}$, we move it from $R_t$ to $A_t$. Here $\tilde{h} = \lceil\zeta h\rceil$, and $\zeta >0$. Algorithm~\ref{alg:add} summarizes the implementation of the ADD operation.

\begin{algorithm}[H]
	\KwData{ $\theta_t$, $r_t$, $R_t$, $A_t$, $X$ }
	\KwResult{ $R_{t+1}$, $A_{t+1}$  }
	Set  $h = \lceil c\log(\frac{md+mx}{\lambda})\log(p) \rceil$\;
	$\tilde{h} = \lceil \zeta h \rceil$ \;
	
	\For{$l = 1$ \KwTo $h$  }
	{
		$i \leftarrow \max_{\hat{i}\in R_t} |x_{\hat{i}}^T \theta_t|$ \;
		Set $V_i = \{ \hat{i}\big| \hat{i} \in \mathcal{R}_t, \hat{i} \neq i, \big||x_{i}^T \theta_t| - ||x_{i}||_2 r_t\big| \leq |x_{\hat{i}}^T \theta_t|+||x_{\hat{i}}||_2r_t\}$ \;
		
		\eIf{  $|V_i| < \tilde{h} $ }
		{ $A_{t} \leftarrow A_{t} \cup \{i\} $ \;
			$R_{t} \leftarrow R_{t} - \{i\} $ \;
		}{Stop\;}
	}
	$A_{t+1} \leftarrow A_{t} $ \;
	$R_{t+1} \leftarrow R_{t} $ \;
	\caption{Algorithm for ADD operation}\label{alg:add}
\end{algorithm}

\section{Convergence Analysis}

In this section, we first discuss the convergence properties of SAIF and then provide the detailed complexity analysis of our SAIF algorithm.  In this manuscript we present the complexity of SAIF with coordinate minimization (CM) as the inner base algorithm.
We can derive the corresponding complexity in a similar way if  an alternative base algorithm such as FISTA~\citep{Beck2009} is employed.

\subsection{Algorithm Properties}
Similar to dynamic screening, SAIF employs coordinate minimization (CM) in the primal variable space. Besides feature screening (by DEL operations), SAIF has feature recruiting operation (ADD). In this subsection, we first discuss the convergence of the base algorithm, then we show that the numbers of  DEL and ADD operations are finite in SAIF. 

\subsubsection{Coordinate Minimization (CM)}

The base algorithm we employ in SAIF is shooting algorithm~\citep{Fu98}, which is a cyclic block coordinate minimization method. Coordinate descent (CD) and coordinate minimization (CM) methods have been studied extensively by many researchers~\citep{NESTEROV2012,Beck2013,Wright2015}. Recently \cite{Li2017} have achieved faster convergence rates for CD and CM methods on convex problems. Based on the analysis from~\cite{Li2017}, we can prove the following lemma regarding CM for LASSO. \footnote{We use $k$ to indicate the iteration or the number of  base operations in CM, and $t$ for the iteration number in the outer loop of SAIF or dynamic screening.}

\noindent\textbf{Lemma 1 (Adaptation of~\cite{Li2017})} {\it For the LASSO problem~\eqref{eq:plasso} with a $\gamma$-convex loss function, at most $\log_{\psi} \frac{\varepsilon}{P(\beta_{0}) - P(\beta^*)} $ base operations are performed  with cyclic coordinate minimization to arrive at  $\beta_a$ such that  $P(\beta_{a}) - P(\beta^*)\leq \varepsilon$, where $\psi =  \frac{p\bar{L}^2}{p\bar{L}^2 + \gamma^2}$, $\bar{L} = \sqrt{\sigma_{\max}}L$, $\sigma_{\max}$ is the largest eigenvalue of $X^TX$, $L$ is the Lipschitz constant of $\mathbf{f}'$, and $\beta_0$ denotes the starting point of the primal variables. }

The base operation (soft-thresholding) in CM is operated in the primal variable space. Feature screening or feature selection operations such as ADD and DEL operations rely on the dual variable estimation. We provide the following lemma to show that the accuracy of dual variable  estimates is almost linearly bounded by the accuracy of primal variable  estimates when the iteration number is large. 

\noindent\textbf{Lemma 2} {\it For the primal problem~\eqref{eq:plasso} and dual problem~\eqref{eq:dlasso}, let $\hat{\theta}_k = -\frac{\mathbf{f}'(X\beta_k)}{ \lambda}$,  $\tau_k =\frac{1}{\max_i |x_i^T\hat{\theta}_k|}$, and $\theta_k = \tau_k \hat{\theta}_k$, with a large enough $k$ in coordinate minimization,  we have $ ||\theta_k - \theta^* ||^2_2 \leq \frac{LM}{\lambda^2} || \beta_k - \beta^* ||^2_{\Sigma}$, where $\Sigma = X^TX $, and $M$ is a constant value.}

The proofs of Lemmas 1 and 2 can be found in Appendix B. With Lemma 2, we can see that the estimation of dual variables relies on the accuracy of primal variable  estimates. In SAIF, the starting point for each $\beta_t$ is already with relatively high accuracy as empirically there are only one or a few features different between steps $t$ and $t-1$.

\subsubsection{Finite number of ADD and DEL Operations}
With CM as the inner base algorithm, we prove that the outer loop can stop in a finite number of steps. The ADD operation recruits more features into the active set, and thus results in decreasing optimal objective value as shown in Theorem 1. Since the DEL operation does not change the optimal objective function value, the corresponding optimal dual objective function value of the sub-problem decreases monotonically and finally converges to the value of the original full-scale problem. Experimentally, for a given $\lambda$, the running time of SAIF is proportional to the size of the optimal active set $\bar{\mathcal{A}}$. The following theorem provides the guarantee for the convergence of SAIF. 

\noindent\textbf{Theorem 3} {\it
Let $\beta_{t}^*$ and  $\theta_{t}^*$ be the optimal primal and dual solutions for the sub-problem with the active feature set $\mathcal{A}_t$. 

a) If $\bar{ \mathcal{A}} \nsubseteq \mathcal{A}_t$,   and $t < t'$, then $ \mathcal{A}_t \neq  \mathcal{A}_{t'}$.

b) $\lim_{t \to \infty}\theta_{t}^* = \theta^* $; $\lim_{t \to \infty}\beta_{t}^* \in S_{ \mathcal{F}}^* $ .

c)  $\exists T, \ \forall t \geq T, \theta_{t}^* = \theta^* $, and $\beta_{t}^* \in S_{F}^* $ . }

\noindent\textbf{Proof}:
a) If $\bar{ \mathcal{A}} \nsubseteq \mathcal{A}_t$,  from Remark 2 in Section 2.1, we can see that, $\exists i \in  \mathcal{R}_t, |x_i^T \theta_t^*| \geq 1$. If $\max_{i \in R_t}, |x_i^T \theta_t^*| >1$, we can apply the ADD operation at step $t$ to add the most active feature to $\mathcal{A}_{t+1}$. We will have $D(\theta^*_t) > D(\theta_{t+1}^*)$. As $t<t'$, $D(\theta^*_t) >  D(\theta_{t+1}^*) \geq D(\theta_{t'}^*)$, and $\mathcal{A}_t \neq A_{t'}$.
If $\max_{i \in R_t}, |x_i^T \theta_t^*| = 1$, the optimal dual variable is already on the hyperplanes $|x_i^T \theta_t^*|$ = 1. From the SAIF algorithm, we can see that, with an ADD operation to move all $x_i:  |x_i^T \theta_t^*| = 1$ to $\mathcal{A}_t$,  the  optimal dual solution will remain the same, i.e., $\theta_{t+1}^* = \theta_{t}^*$. The ADD operation will stop at step $t$, as  $\max_{i \in R_{t+1}}, |x_i^T \theta_{t+1}^*| < 1$. DEL does not remove $x_i:  |x_i^T \theta_t^*| = 1$ from the active set $\mathcal{A}_{t'}, \forall t'>t$, as the optimal dual variable will remain the same, and the algorithm will stop. Thus  $\mathcal{A}_t \neq \mathcal{A}_{t'}$. In summary, we have  $\mathcal{A}_t \neq \mathcal{A}_{t'}, \forall t' >t$.

b) At step $t$, if the operation is DEL, we have $P(\beta^*_t) = P(\beta^*_{t+1})$, and $D(\theta^*_t) = D(\theta^*_{t+1})$, as removing inactive features does not change the primal and dual problems. If the operation is ADD, and  $\max_{i \in \mathcal{R}_t} |x_i^T \theta_t^*| > 1$, we have  $P(\beta^*_t) > P(\beta^*_{t+1})$, and $D(\theta^*_t) > D(\theta^*_{t+1})$. Thus $\exists m>0, D(\theta_t^*) > D(\theta_{t+m}^*)$ for each step $t$, which means $D(\theta_t^*)$ will converge to a fixed value as $t \to \infty$. From a), $\mathcal{A}_{t}$ changes monotonously with a finite number of combinations. Thus SAIF will stop within a finite number of steps. Let $\lim_{t \to \infty} D(\theta_{t}^*) = \bar{d}$, and $ \Gamma = \{\theta | D(\theta) = \bar{d}, \ \theta \in \lim_{t \to \infty} \Omega_{\mathcal{A}_t}\}$. As $\Omega_{\mathcal{A}_t} \supseteq \Omega_{\mathcal{F}}$, we have $\bar{d} \geq D(\theta^*) $. If $\theta^* \notin \Gamma$,  as the dual objective function is smooth and convex, and  $\Omega_{\mathcal{F}} \subseteq  \lim_{t \to \infty} \Omega_{\mathcal{A}_t}$, $\forall \hat{\theta}^* \in \Gamma, D(\hat{\theta}^*) = \bar{d} > D(\theta^*)$. As $\theta^* = argmax_{\theta \in \Omega_{\mathcal{F}}}D(\theta)$, and $\theta^*$ is unique, we have $ \forall \hat{\theta}^* \in \Gamma,  \hat{\theta}^* \notin \Omega_{\mathcal{F}}$. This implies $\forall \hat{\theta}^* \in \Gamma, \exists i, |x_i^T\hat{\theta}^*| > 1$, which contradicts the algorithm stopping criterion. Therefore we have  $\theta^* \in \Gamma$. As the optimal dual value is unique, $\lim_{t \to \infty}\theta_{t}^* = \theta^* $ and $\lim_{t \to \infty}\beta_{t}^* \in S_{\mathcal{F}}^*$.

c)  As $\Omega_{\mathcal{A}_t} = \cap_{i \in \mathcal{A}_t}\{\theta: |x_i^T\theta|\leq 1\}$, the active sets at different iterations are different before the algorithm stops from a). From b), we have $\lim_{t \to \infty}\theta_{t}^* = \theta^* $. There are at most $\big(\sum_{k=0}^{k=n_A-1}\binom{n_\mathcal{A}}{k}\big)\big(\sum_{k=0}^{k=n_\mathcal{R}}\binom{n_\mathcal{R}}{k}\big)$ different potential active sets ($n_\mathcal{A} + n_\mathcal{R} = p, n_\mathcal{A}= |\bar{\mathcal{A}}|$) through the algorithm iterations of upating the current active features. In practice, the number of legitimate active set combinations is much smaller. Thus, $\exists T, \ \forall t \geq T, \theta_{t}^* = \theta^* $, and $\beta_{t}^* \in S_{F}^* $.

\subsection{Complexity Analysis}
For complexity analysis, we split the SAIF algorithm into three phases: feature recruiting, inactive feature deletion, and accuracy pursuing. The inactive feature deletion phase is the same as the feature screening phase in dynamic screening. We first present the complexity analysis for dynamic screening, which is our additional contribution, and then based on that and previous results, we give the detailed complexity analysis for SAIF.  

\subsubsection{Complexity Analysis for Dynamic Screening}
Dynamic screening~\citep{GAP, Fercoq2015} starts its active set with the whole feature set.  
Let $r_i$ be the radius of the ball region for the screening of feature $i$,  according to the same screening rule as the DEL operation,  
\begin{align}
&| x_i^T\theta_t| + ||x_i||_2r_i <1
\implies   r_i <\frac{1- \frac{|x_i^T\hat{\theta}_t|}{\max_i|x_i^T\hat{\theta}_t|}}{||x_i||_2} 
= \frac{1- \frac{|x_i^T\hat{\theta}_t|}{|x_m^T\hat{\theta}_t|}}{||x_i||_2} .
\end{align}
Here $m$ is the feature with the value of $\max_i|x_i^T\hat{\theta}_t|$,  $x_m$ is the corresponding feature vector,  $\hat{\theta}_t=-\frac{\mathbf{f}'(X\beta_t)}{\lambda}$, $\theta_t =\tau \hat{\theta}_t $, and $\tau = \frac{1}{\max_i|x_i^T\hat{\theta}_t|}$. If feature $i$ does not belong to the final active set $\bar{ \mathcal{A}}$, then $|x_i^T\theta^*|<1$. With large $t$, $m$ belongs to $\bar{ \mathcal{A}}$ according to Theorem 1, and $|x_m^T\theta^*|=1$. 
We have 

\begin{align}
& r_i < \frac{1- \frac{|x_i^T\hat{\theta}_t|}{|x_m^T\hat{\theta}_t|}}{||x_i||_2} \approx \frac{1-|x_i^T\theta^*|}{||x_i||_2} .
\end{align}
Thus the screening radius for feature $i$ is determined by how close $\hat{\theta}_t$ and $\theta^*$ are, linearly determined by the primal variable accuracy according to  Lemma 2. With $\varepsilon$ as the pre-specified objective function value accuracy, the following theorem gives the time complexity of the dynamic screening procedure.

\noindent\textbf{Theorem 4} {\it Assuming that the time complexity for one coordinate minimization operation is $O(u)$, the time complexity for dynamic screening  is  $O\Big(u\frac{\bar{L}^2}{\gamma^2}\big( p\log\frac{G_0}{\varepsilon_D} + |\bar{ \mathcal{A}}| \log \frac{\varepsilon_D}{\varepsilon} \big)\Big)$. Here $G_0 = P(\beta_0) - P(\beta^*)$, and $\varepsilon_D$ is the  accuracy of the objective function value for the last feature screening operation. }

\noindent\textbf{Remark 3} {\it With coordinate minimization, the number of iterations to reach the accuracy of the objective function value $\epsilon$ is $O\Big(\frac{\bar{L}^2}{\gamma^2}\big(  p\log\frac{1}{\varepsilon_D} + |\bar{ \mathcal{A}}| \log \frac{\varepsilon_D}{\varepsilon} \big)\Big)$. As $p>>|\bar{ \mathcal{A}}|$, the computation cost in dynamic screening is mainly from the iterations to reach $\epsilon_D$.}

The proof of Theorem 4 can be found in Appendix C. Experiments will confirm the conclusions from Theorem 4 and Remark 3 with the results presented in Section 5. 

\subsubsection{Complexity Analysis for SAIF}
With the complexity analysis for dynamic screening, we now derive the complexity of SAIF and show its advantages over dynamic screening theoretically. SAIF starts the algorithm from the feature recruiting phase. The ADD operation recruit a feature with $\max_{i\in \mathcal{R}_t}|x_i^T\theta_t|$. When $\theta_t$ is close to $\theta_t^*$, we have 
\begin{align} 
&|x_i ^T \theta_t| - ||x_i||_2 r_i  > |x_{\hat{i}}^T \theta_t|+||x_{\hat{i}}||_2r_i , \  \forall \hat{i} \in \mathcal{R}_t, \hat{i}\neq i \\
\implies &r_i < \frac{|x_i^T\theta_t| - |x_{\hat{i}}^T\theta_t|}{||x_i||_2 + ||x_{\hat{i}}||_2} \approx \frac{|x_i^T\theta^*_t| - |x_{\hat{i}}^T\theta^*_t|}{||x_i||_2 + ||x_{\hat{i}}||_2} \  \forall \hat{i} \in \mathcal{R}_t, \hat{i}\neq i .\label{eq:addr}
\end{align}
Here we use $\theta^*_t$ rather than $\theta^*$ as the algorithm has not reached the stopping point of ADD operations and $\bar{ \mathcal{A}} \nsubseteq \mathcal{A}_t$. In~\eqref{eq:addr},  the radius for adding feature $i$ into the active set is determined by how much it can outperform the other features. 
 We use $T_a$ to represent the running time consumed in the feature recruiting phase. The inactive feature deletion phase starts from setting IsADD = False in SAIF in Algorithm~\ref{alg:saif}. Let $p_A$ be the total number of features involved in the ADD operation; after the $d$th feature ($d$ in the sequence of $\{1,2,...,d,...,p_A\}$) has been added into the active set, we use $P_d$, $p_d$, and $\beta^*_d$  to denote the primal objective function, the size of the active set,  and the optimal primal solution of the sub-problem, respectively. Let $Q_d(\beta) = P_d(\beta) - P_d(\beta^*_d)$. The time complexity for SAIF with CM  is given by the  following lemma and theorem.

\noindent\textbf{Lemma 3} {\it With $O(u)$ as the complexity for the base operation of coordinate minimization of the LASSO problem with a $\gamma$-convex loss function, $p_A$ the total number of features involved in ADD operations, and $p_{p_A}$ the size of the active set when IsADD is set to false, the complexity for the feature recruiting phase in SAIF is 
	\begin{align} \nonumber
	\frac{Ku+pn}{K} \Big( \Upsilon + \frac{\bar{L}^2}{\gamma^2} \Phi+ p_{p_A} \frac{\bar{L}^2}{\gamma^2} \log \frac{\bar{Q}}{Q_{p_A}(\beta_{p_A})} \Big),
	\end{align} 
	where 
	\begin{align*} 
	&\bar{Q} = \big(\Pi_{d=1}^{p_A-1}Q_{d+1}(\beta_d)^{p_{d+1}-p_d}\big)^{\frac{1}{p_{A}}} , \ \  \Upsilon = \log\big( \Pi_{d=1}^{p_A -1 } \frac{Q_{d+1}(\beta_d)}{Q_d(\beta_d)^{\frac{p_d}{p_{d+1}}}} \frac{1}{Q_{p_A}(\beta_{p_A})} \big) , \\
	&\Phi =  \log\big( \Pi_{d=1}^{p_A -1} \frac{Q_{d+1}(\beta_d)^{p_{d}}}{Q_d(\beta_d)^{p_d}}\big). 
	\end{align*}
}

\noindent\textbf{Theorem 5} {\it With $O(u)$ as the complexity for the base operation of coordinate minimization of the LASSO problem  with a $\gamma$-convex loss function, 
	the time complexity for SAIF is $O\bigg(u\frac{\bar{L}^2}{\gamma^2} \big( \bar{p} \log \frac{\bar{Q}}{\varepsilon_D} + \bar{p} p_A + |\bar{ \mathcal{A}}|\log \frac{\varepsilon_D}{\varepsilon} \big) \bigg) $. Here $\bar{p}$ is the maximum size of the active set during the algorithm iterations, $\bar{Q}$ is the geometric mean of the accuracies of the sub-problem objective function values corresponding to each ADD operation, and $\varepsilon_D$ is the accuracy of the objective function value for the last feature DEL operation. } 

\noindent\textbf{Remark 4} {\it With coordinate minimization, the number of iterations to reach the accuracy of the objective function value $\epsilon$ is $O\Big(\frac{\bar{L}^2}{\gamma^2}\big(  \bar{p}\log\frac{\bar{Q}}{\varepsilon_D}  + \bar{p} p_A + |\bar{ \mathcal{A}}| \log \frac{\varepsilon_D}{\varepsilon} \big)\Big)$.  $\bar{Q}$ is a value much smaller than $G_0$ in dynamic screening (as the value of $Q_d$ for adding feature $d$ usually is very small). }

The proofs for Lemma 3 and Theorem 5 are given in Appendix C. According to our experiments, $\bar{p}$ is often close to the number of the actual active features in the optimal LASSO solution, $|\bar{\mathcal{A}}|$. The dominating factor for the computational complexity of SAIF is the second term $\bar{p} p_A$. The fewer features being added in the active set, the less time SAIF will consume. Experimentally,  $p_A$ is often a value several times larger than $|\bar{\mathcal{A}}|$, and $p_A << p$.   We can conclude that SAIF takes much less time than dynamic screening based on the analysis of Theorems 4 and 5. With the theoretical safe and convergence guarantees, SAIF can work with extremely high-dimensional data to obtain optimal LASSO solutions. 

\section{SAIF for Fused LASSO}
The active incremental philosophy can be applied to many convex sparse models. In this section, we show how to scale up fused LASSO solutions based on SAIF. 
The formulation for fused LASSO is 

\begin{equation}
\  \  \min_{\beta}\sum_{j=1} ^n f(x_{j\bullet} \beta, y_j) + \lambda ||D\beta||_1,  \label{eq:genls}
\end{equation}
where $||D\beta||_1 = \sum_{(a,b) \in E} |\beta_a - \beta_b|$, and each pair in $E$ denotes an edge in a  tree with $\mathcal{F}$ as the vertex set. The tree $G(\mathcal{F},E)$ captures the dependency structures among features.  Here $D$ is a matrix representation of the tree, and in each row of $D$, we have zero entries except two with values equal to $1$ and $-1$, respectively.  With $M_{-p}$ denotes the reduced matrix of a given matrix (vector) $M$ without its corresponding $p$th column (entry), the fused LASSO problem can be further transformed into the equivalent LASSO formulation with the following theorem.

\noindent\textbf{Theorem 6} {\it If $D$ can be transformed into a diagonal matrix with a column transformation matrix T, i.e. $\tilde{D} = D T$, and $\tilde{D}$ is a diagonal matrix, then

a) the problem~\eqref{eq:genls} is equivalent to    
\begin{align}
& \tilde{P}:\  \  \min_{\tilde{\beta}, b}\sum_{j=1}^n f\Big(\sum_{i=1}^{p-1} \tilde{x}_{ji} \tilde{\beta}_i + \tilde{x}_{jp} b, y_j\Big) + \lambda ||\tilde{\beta}||_1, \label{eq:fusedtp}
\end{align}

where $\tilde{X} = XT$, and the solution relationship is $\beta^* = T \bigl[\begin{smallmatrix}
\tilde{\beta}^* \\b^*
\end{smallmatrix} \bigr]$; 

b) a dual form of~\eqref{eq:fusedtp} is 
\begin{align}
&\tilde{D}: \ \  \min_{\bar{\theta} \in \Omega} -\sum_{j=1}^{n} f^*(-\lambda \bar{\theta}_j), \,\, \Omega = \big\{\bar{\theta}:  |\bar{x}_i^T\bar{\theta}|  \leq 1, \forall i \in \{1,...,p-1\} \big\}. \label{eq:fusedtd}
\end{align}

Here $\bar{X} = \tilde{X}_{-p}$, and $H =\left[
\begin{array}{c}
I \\
h 
\end{array}
\right]  $, $ h = \big[-\frac{\bar{x}_{1,p}}{\bar{x}_{n,p}}, ...,-\frac{\bar{x}_{n-1,p}}{\bar{x}_{n,p}}\big]$.   $\bar{\theta} = H\theta_{-p} $ , and $\theta = -\frac{\mathbf{f}'\big(\tilde{X}\bigl[\begin{smallmatrix}
\tilde{\beta}^* \\b^* \end{smallmatrix}\bigr]\big)}{\lambda}$;

c) $\lambda_{max} = \max_{i \in \{1,...,p-1\}} \big|\bar{x}_i^T\mathbf{f}'(\tilde{X}\bigl[\begin{smallmatrix}
\mathbf{0} \\b \end{smallmatrix}\bigr])\big|$, and $\bar{x}_i$ is the $i$th column of $\bar{X}$. 
}

With the primal form~\eqref{eq:fusedtp} and dual form~\eqref{eq:fusedtd} in Theorem 6, we just need a transformation on the feature set to apply SAIF to fused LASSO problems. From the proof of Theorem 2 in~\cite{GAP}, we can easily get  $\forall \bar{\theta} \in \Omega,\  \bar{\beta} =  \bigl[\begin{smallmatrix}
\tilde{\beta} \\b \end{smallmatrix}\bigr] \in R^{ p \times 1}, \  \   ||\bar{\theta}^* -\bar{\theta} ||_2^2 \leq \frac{2}{\lambda^2} \bigg[    \tilde{P}(\bar{\beta}) - \tilde{D}(\bar{\theta}) \bigg]. $ With the duality gap, we can derive the ADD and DEL operation for fused LASSO. The following Theorem shows how to project the current dual estimation $\hat{\bar{\theta}}$ in~\eqref{eq:fusedtd} to the feasible space $\Omega$ for regression with the least square loss function.

\noindent\textbf{Theorem 7} \ {\it For linear regression problems with fused LASSO regularization,  $\hat{\bar{\theta}} = \tau \bar{\theta}$ is the scaled feasible dual variable vector  of $\bar{\theta}$ closest to $\bar{\theta}^*$ in~\eqref{eq:fusedtd} with
$\tau=min\big\{max\{\frac{\mathbf{y}^T\bar{\theta}}{\lambda ||\bar{\theta}||_2^2}, \\ -\frac{1}{||\bar{X}^T\bar{\theta}||_{\infty}}\}, \frac{1}{||\bar{X}^T\bar{\theta}||_{\infty}}\big\}$. }

The proofs for Theorems 6 and 7 are presented in Appendix D. The algorithm for fused LASSO is the same as  what we presented for SAIF for LASSO with the transformation steps. As the transformation matrix is highly sparse and only has column operations on the feature matrix $X$, we can replace matrix multiplication with column operations to further improve computation efficiency.

\section{Experiments}

In this section, we present the  experiments comparing SAIF with other existing LASSO and fused LASSO  methods. We first evaluate the selected methods for the LASSO formulation based on a simulation study and then apply them to one real-world study. In the second subsection, we evaluate SAIF for logistic regression with two real-world data sets. We  present the comparison between SAIF and sequential screening as well as homotopy methods in the third subsection. In the last subsection, two studies are performed to evaluate the selected methods for fused LASSO. The base algorithm (coordinate minimization) is implemented with C, and the main algorithms of SAIF, dynamic screening~\citep{GAP}, DPP~\citep{WangLasso}, and the homotopy method~\citep{Zhao2017} are coded with Matlab. We use the provided package~\citep{BLITZ} for the BLITZ method. The experiments are run on an iMac with OS Sierra version 10.12.1 and Intel Core i5. The implementation  environment is the same for all the experiments unless specified.

\subsection{Results for Linear Regression}

Similar to sequential and dynamic screening algorithms, SAIF can be assembled with different kinds of LASSO solution methods. Shooting algorithm (CM) is chosen as the base algorithm in our experiments. Both dynamic screening ~\citep{GAP} and SAIF can do feature screening or selection without the help from a heavier parameter solution. We specifically focus on the performance comparison among (1) shooting algorithm without screening~(No Scr.), (2) shooting algorithm with dynamic screening~\citep{GAP}~(Dyn. Scr), (3) working set method~\citep{BLITZ} (BLITZ), and (4) shooting algorithm with SAIF screening~(SAIF).  All of these are safe methods for LASSO problems. 

\subsubsection{Simulation Study}

\begin{figure}
	\centering
	\includegraphics[width=5.6in, height= 1.6in]{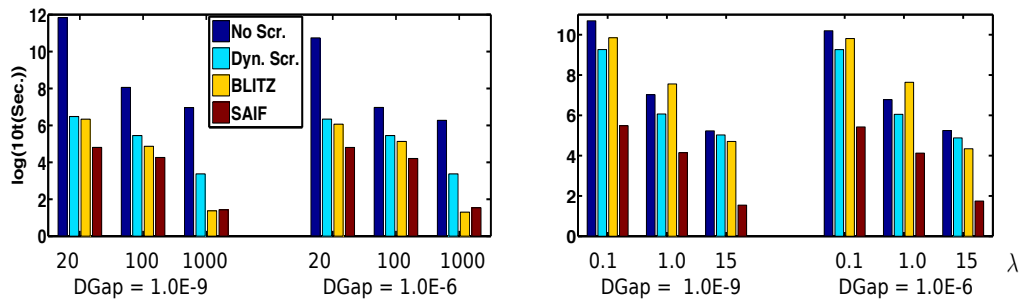}
	\caption{Running time comparison on simulation (left) and breast cancer (right).} 
	\label{fig:bar}
\end{figure}

First, we simulate the data sets with $n= 100$ samples and $p= 5,000$ features according to a linear model $\mathbf{y} = X\beta + \epsilon$, where each column of $X$ is a vector with random values uniformly sampled from the interval $[-10,10]$, and the white noise $\epsilon \sim N(0, 1)$. For the linear coefficients $\beta$, $20\%$ entries ($0.2p$) are randomly set to the values in $[1,-1]$, and the rest ($0.8p$) to zero. For this data set, we can derive $\lambda_{max}= 2.183\text{E}4$.  The left plot in Figure~\ref{fig:bar} illustrates the running time for different methods in the logarithmic time scale with $\lambda = 20$, $100$, and $1,000$  at different stopping accuracy $1.0\text{E}-9$ and $1.0\text{E}-6$  for the desired duality gap (DGap). We can see that, SAIF takes much less time than the other methods to reach the optimal solutions with the specified accuracy. The results also show that SAIF is more efficient  compared to the existing safe methods when $\lambda$ is small.  

\subsubsection{Breast Cancer Data}

Breast cancer data set consists of gene expression data of 8,141 genes for 78 metastatic and 217 non-metastatic breast cancer patients from the study introduced in~\cite{Chuang2007}. In this set of experiments, the metastatic samples are labeled as 1 and non-metastatic as -1 as the output of the LASSO linear regression problem. The right plot in Figure~\ref{fig:bar} compares the running time for the same four different methods at different $\lambda$'s. Again, SAIF takes the least computation time at different duality gaps.

\begin{figure}[b!]
	\centering
		\vspace{-0.1in}
	\includegraphics[width=5.5in, height= 1.5in]{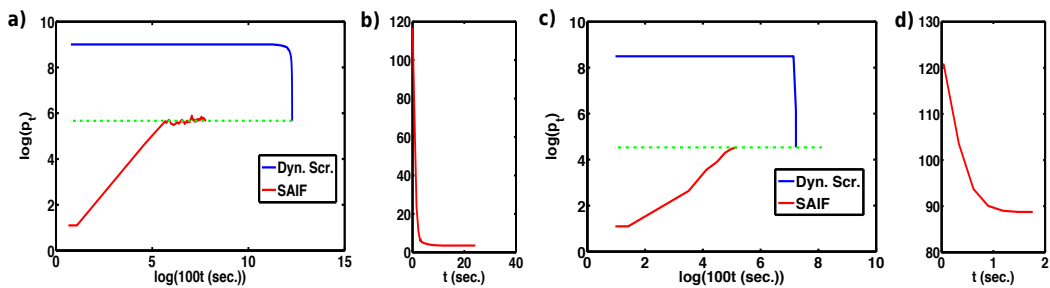}
	\caption{a,c): Active feature set size at different time points for breast cancer data with $\lambda = 0.1$ and $5$, respectively. Green dotted lines indicate the optimal feature set size. b,d): The corresponding $D(\theta_t)$ value changes with different time points during SAIF optimization at $\lambda = 0.1$ and $5$, respectively. } 
	\label{fig:bcancer_feat_dual}
\end{figure}

We further investigate the size of the active set along with the optimization iterations for dynamic screening and SAIF in  Figures~\ref{fig:bcancer_feat_dual}-a,c), with $\lambda = 0.1$ and $5$. We can see that SAIF starts from a small active feature set and gradually increases its size with time, while dynamic screening starts from the whole feature set and takes longer time to reach the point with screening power. Figures~\ref{fig:bcancer_feat_dual}-c,d) illustrate the change of the dual objective function values $D(\theta_t)$ for SAIF during the optimization procedure. With the active feature set size increasing, $D(\theta_t)$ decreases and finally converges to a steady value $D(\theta^*)$, indicating the algorithm obtains the optimal solutions to the original LASSO problems. 

\begin{figure}
	\centering
	\includegraphics[width=5.5in, height= 3.0in]{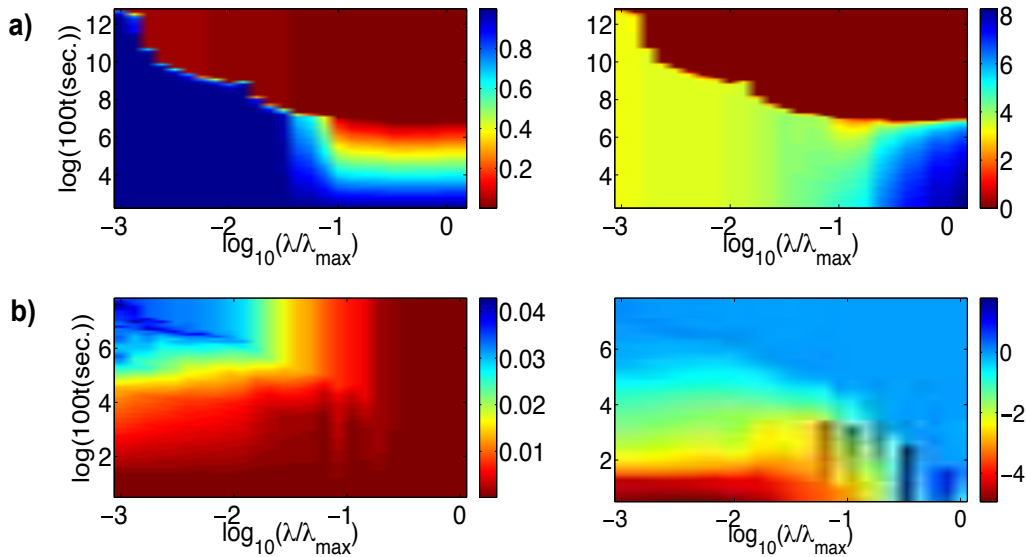}
	\caption{ $\frac{p_t}{p}$ (left) and  $\log(\frac{p_t}{p'})$ (right) as functions of $\log_{10}\frac{\lambda}{\lambda_{max}}$ (x-axis) and $\log(100\times t(sec.))$ (y-axis) for dynamic screening (a), and SAIF (b) on breast cancer data.}     
	\label{fig:bcancerP}
\end{figure}

Let $p_t$ be the feature number at step $t$ for SAIF or dynamic screening. The left column in  Figure~\ref{fig:bcancerP} shows the change of $\frac{p_t}{p}$ with respect to the regularization penalty ($\log_{10}\big(\frac{\lambda}{\lambda_{max}}\big)$ on x-axis) and the optimization time ($\log(100\times t(sec.))$ on y-axis). Similarly, we plot the change of $\log(\frac{p_t}{p'})$, where $p'$ is the corresponding optimal active feature size in the right column of Figure~\ref{fig:bcancerP}. From Figure~\ref{fig:bcancerP}, it is clear that dynamic screening always takes longer time to reach the optimal active feature set size, especially when $\lambda$ is small. Before reaching the point with screening power, the active feature set size is almost $p$. While the active feature set size for SAIF grows gradually from a small set. Due to the small active set size for the starting iterations, SAIF can more efficiently reach the optimal active set size with much shorter running time. All of these results confirms the theoretical complexity analysis for dynamic screening and SAIF. Furthermore, both Figures~\ref{fig:bcancer_feat_dual} and~\ref{fig:bcancerP} illustrate that SAIF is more scalable than the existing methods as it always starts from a very small active set and iteratively focuses on a small subset of the features.

\subsection{Results for Logistic Regression}

\begin{figure}
	\centering
	\includegraphics[width=5.6in, height= 1.7in]{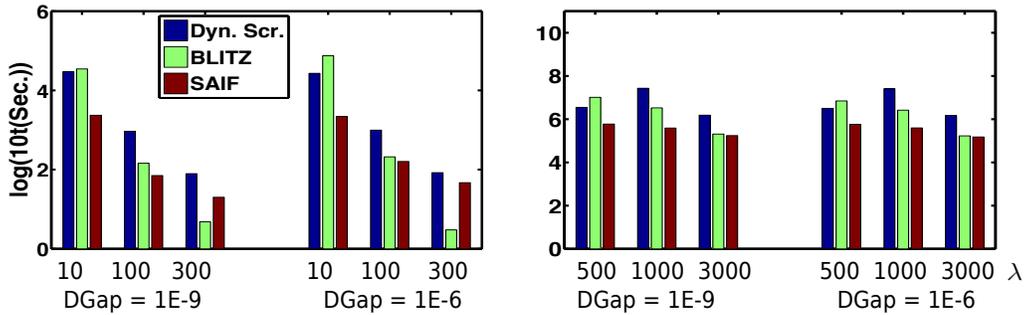}
	\caption{Running time comparison on USPS (left) and Gisette (right) data sets. } 
	\label{fig:logbar}
\end{figure}

We evaluate the proposed algorithms for sparse logistic regression with two data sets, Gisette and USPS, from LibSVM~\citep{Chang2011} Website. The Gisette data set has 5,000 features and 6,000 samples. There are 256 features, 7,291 samples, and 10 labels in the USPS data set; and we categorize the label values large  than 4 as positive, and negative otherwise. For these two data sets, $\lambda_{max}$ is 932,575 and 992, respectively. We use the logistic regression code from L1General~\citep{L1General} as the inner iteration algorithm in this set of experiments.  Figure~\ref{fig:logbar} plots the running time at different $\lambda$ values for dynamic screening, BLITZ, and SAIF. Although BLITZ may achieve comparable performance in a couple of cases when the active set is very small, SAIF consistently takes less computation at different $\lambda$ values for both data sets. From all these results, SAIF can achieve more computational efficiency for both linear and logistic regression compared to the existing safe screening methods. 

\subsection{Comparison with Sequential Screening and Homotopy Methods}
\begin{figure}
	\centering
	\includegraphics[width=5.6in, height= 1.9in]{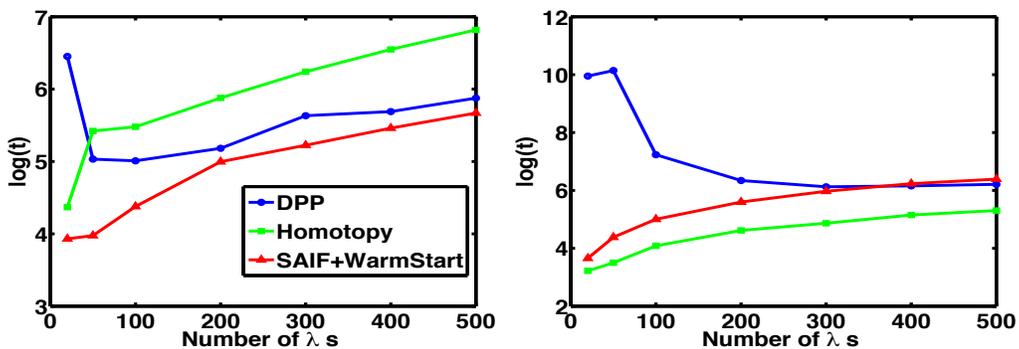}
	\caption{ Running time for different algorithms with different numbers of $\lambda$ values on simulation (left) and breast cancer (right) data sets. } 
	\label{fig:seq}
\end{figure}

With a sequence of decreasing $\lambda$ values,  SAIF can be further improved  with the warm start strategy. Given the simulation and the breast cancer data sets in Section 5.1, a decreasing sequence of $\lambda$ values are evenly sampled from the logarithmic scale of the  range  $[0.001\lambda_{max}, \lambda_{max}]$.  The plots in Figure~\ref{fig:seq} present the running time for DPP~\citep{WangLasso}, the homotopy method~\citep{Zhao2017}, and SAIF with a different number of $\lambda$ values on both data sets. In this set of experiments, we set the stopping criterion with the duality gap $1.0\text{E}-6$ for all the algorithms for fair comparison. The results show that SAIF takes much less time than the DPP method especially when the number of $\lambda$ values is small. With breast cancer data set, the homotopy method can achieve the least computational cost; however, in the result for simulation data, the homotopy method loses its advantages. More critically, the homotopy methods do not have the safe guarantee. Table~\ref{tb:act_feat_hom} provides the average (Avg.) and standard derivation (Std.) for recall (Rec.) and precision (Prec.) regarding the active features recovered by the homotopy method~\citep{Zhao2017}. According to the recall values, the homotopy method always misses some of the active features at different numbers of $\lambda$ values. Furthermore, the homotopy method leads to the inclusion of inactive features  into the final solution as evidenced in Table 1 that the precision cannot reach 1 at different numbers of $\lambda$ values. On the contrary, our SAIF has the safe guarantee: the recall and precision values regarding active features recovered by SAIF are always one.  Clearly, the unsafe strategies employed by the homotopy method do not always reduce computation, and the employed inactive features may lead to larger CPU time consumption as shown in the left plot in Figure~\ref{fig:seq}.

\begin{table*}
	\centering
	\caption{ Recall and precision for active features recovered by the homotopy method~\citep{Zhao2017}. }  \label{tb:act_feat_hom} 
	\begin{center}
		\begin{tabular}{ c|c|c |c | c }
			\hline
		 Num. of $\lambda$ values &Rec. Avg.&Rec. Std &Prec.  Avg. &Prec.  Std \\ \hline
		 20  &0.896  & 0.097 &0.972 & 0.032  \\  \hline
		 50  &0.912 & 0.075 &0.982 &0.017  \\ \hline
		100  &0.911  & 0.079  & 0.979 & 0.021 \\ \hline
		200  &0.926 & 0.061& 0.974 & 0.068 \\ \hline
		300  &0.927 & 0.060 &0.969 & 0.093  \\  \hline
		400  &0.929 & 0.059& 0.971 & 0.087  \\ \hline
		500  &0.929  & 0.058  & 0.976 & 0.060 \\ \hline
		\end{tabular}
	\end{center}
\end{table*}

\subsection{Results for Fused LASSO}

We further present the experiments for fused LASSO with the formulation~\eqref{eq:genls}. There are a few solvers that are suitable for tree fused LASSO problems, such as ~\cite{CVX} and the path solution method~\citep{Arnold2016}. Due to the scalability and solution accuracy issues with the path solution package, we only take~\cite{CVX} as the baseline for comparison in our experiments. We first compare the running time between SAIF and~\cite{CVX} on breast cancer data regarding fused LASSO linear regression; then we  compare them on the FDG-PET data set (ADNI) by logistic regression.

\subsubsection{Breast Cancer Data} 
\begin{figure}
	\centering
	\includegraphics[width=5in, height= 1.6in]{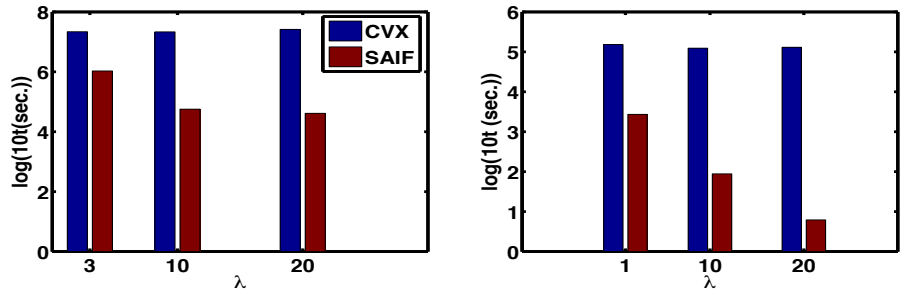}
	\caption{Running time for fused LASSO on breast cancer (left) and PET (right) data sets at duality gap $1.0\text{E}-6$. } 
	\label{fig:fusedlasso}
\end{figure}

For the same breast cancer data set as in the previous subsections, we would like to incorporate the interaction relationships among genes to formulate the fused LASSO problems for regression analysis. 
The largest connected component in the human protein-protein interaction (PPI) network was identified in~\cite{Chuang2007} to capture the gene-gene relationships by a connected graph with 7,782 nodes. The left plot in Figure~\ref{fig:fusedlasso} gives the running time for both CVX and SAIF at different $\lambda$'s for the specified duality gap $1.0\text{E}-6$. The results show that SAIF can significantly reduce computation cost compared to the CVX solver without screening for fused LASSO. 

\subsubsection{FDG-PET Data Set}
The FDG-PET data set has 74 Alzheimer's disease~(AD) patients, 172 mild cognitive impairment~(MCI) subjects, and 81 normal control~(NC) subjects, downloaded from the Alzheimer's Disease Neuroimaging Initiative (ADNI) database. 116 features (each feature corresponds to a brain region) can be derived for each subject after preprocessing. We further use the method described in~\cite{Yang2012} to construct a correlation tree on these features. We take AD as positive (1) and NC as negative (0), and all of MCI samples are not used in this set of experiments in fitting to a fused LASSO logistic regression model. The right plot in Figure~\ref{fig:fusedlasso} gives the running time for three $\lambda$ values at duality gap $1.0\text{E}-6$. Again SAIF  takes much less time than CVX does on this data set.

\section{Conclusions}
In this paper, we have developed a novel feature selection method for LASSO---SAIF. From the experimental results, SAIF can achieve improved efficiency compared with existing methods. SAIF has the potential to scale up for data sets with high dimensional features due to its incremental property. Theoretical analysis reveals  the safety guarantee and low algorithm complexity of the proposed method. Furthermore, SAIF can be potentially extended to group LASSO~\citep{GLasso} and other sparse models. SAIF provides us with a new direction for scaling up sparse learning. Given a data set with extremely high feature dimension, SAIF can be further improved with the multi-level active set and remaining set schema.

\section*{Appendix A. Proof of Theorem 2}
\noindent\textbf{Theorem 2} {\it For the LASSO problem~\eqref{eq:plasso} with the loss function $\mathbf{f}$, if $\mathbf{f}^*$ is $\frac{1}{\alpha}$-strongly convex,  and $\theta^*_0$ and $\theta^*$ are the optimal solutions to the dual problem~\eqref{eq:dlasso} at $\lambda_0$ and $\lambda$ with $\lambda < \lambda_0$, then 
	\begin{equation*}
	||\theta^* - \frac{\lambda_0}{\lambda}\theta^*_0||_2^2 \leq \frac{2\alpha}{ \lambda^2} \bigg[ \mathbf{f}^*(-\frac{\lambda^2}{\lambda_0}\theta^*_0) - \mathbf{f}^*(-\lambda_0\theta^*_0) + (\lambda -\lambda_0)\langle  \mathbf{f}'^{*}(-\lambda_0\theta^*_0), \theta^*_0 \rangle \bigg]. 
	\end{equation*}
	If we have $\theta \in \Omega$, the bound can be further improved by 
	\begin{equation*}
	||\theta^* - \frac{\lambda_0}{\lambda}\theta^*_0||_2^2 \leq \frac{2\alpha}{ \lambda^2} \bigg[ \mathbf{f}^*(-\lambda \bar{\theta}(\bar{\varrho})) - \mathbf{f}^*(-\lambda_0\theta^*_0) + (\lambda -\lambda_0)\langle  \mathbf{f}'^{*}(-\lambda_0\theta^*_0), \theta^*_0 \rangle \bigg],
	\end{equation*}
	where $\bar{\theta}(\bar{\varrho}) = (1-\bar{\varrho})\theta + \bar{\varrho} \frac{\lambda}{\lambda_0}\theta_0^*$, with $\bar{\varrho} = argmin_{\varrho: 0 \leq \varrho \leq 1} \mathbf{f}^*(-\lambda\bar{\theta}(\varrho))$. }

\noindent\textbf{Proof}: As $f^*$ is $\frac{1}{\alpha}$-strongly convex, we have
\begin{equation*}
||\lambda\theta^* - \lambda_0\theta^*_0||_2^2 \leq 2\alpha \bigg[ \mathbf{f}^*(-\lambda\theta^*) - \mathbf{f}^*(-\lambda_0\theta^*_0)  - \langle \mathbf{f}'^{*}(-\lambda_0\theta^*_0), -\lambda\theta^* - (-\lambda_0 \theta^*_0) \rangle \bigg],   \label{eq:strconv}
\end{equation*}
and therefore
\begin{equation}
||\theta^* - \frac{\lambda_0}{\lambda}\theta^*_0||_2^2 \leq \frac{2\alpha}{ \lambda^2} \bigg[ \mathbf{f}^*(-\lambda\theta^*) - \mathbf{f}^*(-\lambda_0\theta^*_0) + \langle  \mathbf{f}'^{*}(-\lambda_0\theta^*_0), \lambda\theta^* -\lambda_0\theta^*_0 \rangle \bigg].
\label{eq:lamapr1}
\end{equation}

As $\theta_0^*$ is the optimal solution at $\lambda_0$ we can see $\theta_0^* \in \Omega$, and $\frac{\lambda}{\lambda_0}\theta_0^* \in \Omega$, thus
\begin{align}
&\mathbf{f}^*(-\lambda \theta^*) \leq \mathbf{f}^*(-\lambda \frac{\lambda}{\lambda_0}\theta_0^*) \label{eq:lamapr2} .
\end{align}

Also as $\theta_0^*$ is the optimal dual solution at $\lambda_0$, thus we have 
\begin{align}
&\langle -\lambda_0 \mathbf{f}'^*(-\lambda_0 \theta_0^*), \theta^* - \theta^*_0 \rangle \geq 0 \\
\implies &  \langle -\mathbf{f}'^*(-\lambda_0 \theta_0^*), \lambda \theta^* - \lambda\theta^*_0 \rangle \geq 0 \\
\implies &  \langle \mathbf{f}'^*(-\lambda_0 \theta_0^*), \lambda \theta^* \rangle \leq \langle \mathbf{f}'^*(-\lambda_0 \theta_0^*), \lambda \theta^*_0 \rangle \label{eq:lamapr3}.
\end{align}

With~\eqref{eq:lamapr1}--\eqref{eq:lamapr3}, we have 

\begin{equation*}
||\theta^* - \frac{\lambda_0}{\lambda}\theta^*_0||_2^2 \leq \frac{2\alpha}{\lambda^2} \bigg[ \mathbf{f}^*(-\frac{\lambda^2}{\lambda_0}\theta^*_0) - \mathbf{f}^*(-\lambda_0\theta^*_0) + (\lambda -\lambda_0)\langle  \mathbf{f}'^{*}(-\lambda_0\theta^*_0), \theta^*_0 \rangle \bigg].
\end{equation*}
For any $\theta \in \Omega$, we have $\bar{\theta} = (1-\varrho)\theta + \varrho \frac{\lambda}{\lambda_0}\theta_0^* \in \Omega, \text{if} \  0 \leq \varrho \leq 1$. This implies
\begin{align*}
&\mathbf{f}^*(-\lambda \theta^*) \leq \min_{\varrho: 0\leq \varrho \leq 1}\mathbf{f}^*(-\lambda \bar{\theta}(\varrho)) \leq \mathbf{f}^*(-\frac{\lambda^2}{\lambda_0}\theta^*_0).
\end{align*}
So we may improve the bound by
\begin{equation*}
||\theta^* - \frac{\lambda_0}{\lambda}\theta^*_0||_2^2 \leq \frac{2\alpha}{ \lambda^2} \bigg[ \mathbf{f}^*(-\lambda \bar{\theta}(\bar{\varrho})) - \mathbf{f}^*(-\lambda_0\theta^*_0) + (\lambda -\lambda_0)\langle  \mathbf{f}'^{*}(-\lambda_0\theta^*_0), \theta^*_0 \rangle \bigg],
\end{equation*}
where $\bar{\theta}(\bar{\varrho}) = (1-\bar{\varrho})\theta + \bar{\varrho} \frac{\lambda}{\lambda_0}\theta_0^*$, and $\bar{\varrho} = argmin_{\varrho: 0 \leq \varrho \leq 1} \mathbf{f}^*(-\lambda\bar{\theta}(\varrho))$. (Q.E.D.)

\section*{Appendix B. Proof of Lemma 1 and Lemma 2}

\noindent\textbf{Lemma 1 (Adaptation of~\cite{Li2017})} {\it For the LASSO problem~\eqref{eq:plasso} with a $\gamma$-convex loss function, at most $\log_{\psi} \frac{\varepsilon}{P(\beta_{0}) - P(\beta^*)} $ base operations are performed  with cyclic coordinate minimization to arrive at  $\beta_a$ such that  $P(\beta_{a}) - P(\beta^*)\leq \varepsilon$, where $\psi =  \frac{p\bar{L}^2}{p\bar{L}^2 + \gamma^2}$, $\bar{L} = \sqrt{\sigma_{\max}}L$, $\sigma_{\max}$ is the largest eigenvalue of $X^TX$, $L$ is the Lipschitz constant of $\mathbf{f}'$, and $\beta_0$ denotes the starting point of the primal variables. }

\noindent\textbf{Proof:}
With $L$ as the Lipschitz constant of $\mathbf{f}'$, $\bar{L} = \sqrt{\sigma_{\max}}L$ is the Lipschitz constant of $X^T\mathbf{f}'$. Following the proof of Theorem 8 by~\cite{Li2017}, we have 
\begin{align*}
& P(\beta_{k+1}) - P(\beta^*) \leq  \frac{p\bar{L}^2}{2\gamma}||\beta_{k+1} - \beta_k||_2^2 \ .
\end{align*}
We have
\begin{align*}
P(\beta_{k}) - P(\beta^*) &=P(\beta_{k}) -P(\beta_{k+1}) + P(\beta_{k+1}) - P(\beta^*) \\
&\geq  \frac{\gamma}{2}||\beta_k - \beta_{k+1}||_2^2 + P(\beta_{k+1}) - P(\beta^*) \\
& \geq (1+ \frac{\gamma^2}{p\bar{L}^2})\big(P(\beta_{k+1}) - P(\beta^*) \big). 
\end{align*}
Hence,
\begin{align}
\frac{P(\beta_{k+1}) - P(\beta^*)}{P(\beta_{k}) - P(\beta^*)} \leq \frac{p\bar{L}^2}{p\bar{L}^2 + \gamma^2} = \psi \ \label{eq:pdratio}.
\end{align}

Recursively applying~\eqref{eq:pdratio} $k$ times, we have 
\begin{align*}
&\frac{P(\beta_{k}) - P(\beta^*)}{P(\beta_{0}) - P(\beta^*)} \leq \psi^k .
\end{align*}

The algorithm reaches the desired accuracy smaller than $\varepsilon$, and it means $ P(\beta_{k}) - P(\beta^*) \leq \varepsilon$. We can set
\begin{align*}
(P(\beta_{0}) - P(\beta^*))\psi^k \leq \varepsilon .
\end{align*}

 Hence, for any iteration number $k \geq \log_{\psi} \frac{\varepsilon}{P(\beta_{0}) - P(\beta^*)}$, we always have the primal gap $P(\beta_{k}) - P(\beta^*) \leq \varepsilon$. (Q.E.D.) 

\noindent\textbf{Lemma 2} {\it For the primal problem~\eqref{eq:plasso} and dual problem~\eqref{eq:dlasso}, let $\hat{\theta}_k = -\frac{\mathbf{f}'(X\beta_k)}{ \lambda}$,  $\tau_k =\frac{1}{\max_i |x_i^T\hat{\theta}_k|}$, and $\theta_k = \tau_k \hat{\theta}_k$, with a large enough $k$ in coordinate minimization,  we have $ ||\theta_k - \theta^* ||^2_2 \leq \frac{LM}{\lambda^2} || \beta_k - \beta^* ||^2_{\Sigma}$, where $\Sigma = X^TX $, and $M$ is a constant value.}

\noindent\textbf{Proof}:  Let $m$ be the feature achieving $\max_i |x_i^T\hat{\theta}_k|$, $\hat{\theta}_k = \theta^* +\rho_k$,  and we get $\tau_k = \frac{1}{|x_m^T\hat{\theta}_k|}= \frac{1}{|x_m^T\theta^* + x_m^T\rho_k|}$. With $\Sigma = X^TX$, for a vector $\mathbf{v}$, we define $||\mathbf{v}||_{\Sigma} = \mathbf{v}^T X^TX \mathbf{v}$. 
Since
\begin{align*}
\lim_{k \to \infty} ||\rho_k||_2 = & \lim_{k \to \infty}||\hat{\theta}_k - \theta^*||_2  = \lim_{k \to \infty} \frac{1}{\lambda}||\mathbf{f}'(X\mathbf{\beta}_k) - \mathbf{f}'(X\mathbf{\beta^*})||_2 \\
&\leq \lim_{k \to \infty} \frac{L}{\lambda}||\mathbf{\beta}_k - \mathbf{\beta^*}||_{\Sigma} \to 0,
\end{align*}

and  $\forall i\in \bar{ \mathcal{A}}, \ |x_i^T\theta^*| = 1$, and  otherwise  $\forall i\in \mathcal{F} - \bar{ \mathcal{A}}, \ |x_i^T\theta^*| <1$, we always can reach $| x_m^T\rho_k|<1$ and $| x_m^T\theta^*|\leq 1$ for large enough k.  Let $|x_m^T\theta^* + x_m^T\rho_k| = 1+  \varphi_k$, we get $\tau_k = \frac{1}{1+ \varphi_k}$. Here $ -|x_m^T\rho_k| \leq \varphi_k \leq |x_m^T\rho_k|$. With the definition of $\theta_k$, 
\begin{align*}
&||\theta_k - \theta^*||^2_2 
=||\frac{\tau_k\mathbf{f}'(X\mathbf{\beta}_k)}{  \lambda} - \frac{\mathbf{f}'(X\mathbf{\beta^*})}{ \lambda} ||_2^2 
=\frac{1}{\lambda^2} ||\frac{\mathbf{f}'(X\mathbf{\beta}_k)}{ 1+\varphi_k} - \mathbf{f}'(X\mathbf{\beta^*}) ||_2^2 \ .
\end{align*}

With $|\varphi_k|\leq |x_m^T\rho_k|<1$, $\tau_k = \frac{1}{1+ \varphi_k} =\sum_{i=0}^{\infty} (-1)^i \varphi_k^i$. Let $\Phi = \sum_{i=1}^{\infty} (-1)^{i+1}\varphi_k^i$. We have $\tau_k = 1 -  \Phi$, and $\Phi =\tau_k \varphi_k $.
Hence,
\begin{align*}
&||\theta_k - \theta^*||^2_2 \\
=&  \frac{1}{\lambda^2} ||\mathbf{f}'(X\mathbf{\beta}_k)(1 -\Phi ) - \mathbf{f}'(X\mathbf{\beta^*}) ||_2^2 \\
= &  \frac{1}{\lambda^2} \langle (\mathbf{f}'(X\mathbf{\beta}_k) - \mathbf{f}'(X\mathbf{\beta^*}) )- \Phi \mathbf{f}'(X\mathbf{\beta}_k), (\mathbf{f}'(X\mathbf{\beta}_k) - \mathbf{f}'(X\mathbf{\beta^*}) )- \Phi \mathbf{f}'(X\mathbf{\beta}_k) \rangle \\
= &  \frac{1}{\lambda^2}||\mathbf{f}'(X\mathbf{\beta}_k) - \mathbf{f}'(X\mathbf{\beta^*})||_2^2 +  \frac{1}{\lambda^2} || \Phi \mathbf{f}'(X\mathbf{\beta}_k)||_2^2 -  \frac{2}{\lambda^2}  \langle (\mathbf{f}'(X\mathbf{\beta}_k) - \mathbf{f}'(X\mathbf{\beta^*}) ), \Phi \mathbf{f}'(X\mathbf{\beta}_k) \rangle  \label{eq:thetabd} \\
= &\frac{1}{\lambda^2}||\mathbf{f}'(X\mathbf{\beta}_k) - \mathbf{f}'(X\mathbf{\beta^*})||_2^2 + \frac{\Phi^2 }{\lambda^2} || \mathbf{f}'(X\mathbf{\beta}_k)||_2^2 - \frac{2\Phi }{\lambda^2}  \langle (\mathbf{f}'(X\mathbf{\beta}_k) - \mathbf{f}'(X\mathbf{\beta^*}) ),  \mathbf{f}'(X\mathbf{\beta}_k) \rangle \  .
\end{align*}

As $\Phi =\tau_k \varphi_k $, we get
\begin{align} \nonumber
||\theta_k - \theta^*||^2_2 =& \nonumber \frac{1}{\lambda^2}||\mathbf{f}'(X\mathbf{\beta}_k) - \mathbf{f}'(X\mathbf{\beta^*})||_2^2 + \frac{\tau_k^2\varphi_k^2 }{\lambda^2 } || \mathbf{f}'(X\mathbf{\beta}_k)||_2^2  \\ 
&- \frac{2\tau_k\varphi_k}{\lambda^2}  \langle  (\mathbf{f}'(X\mathbf{\beta}_k) - \mathbf{f}'(X\mathbf{\beta^*}) ), \mathbf{f}'(X\mathbf{\beta}_k)  \rangle\\ 
= & \frac{1}{\lambda^2}||\mathbf{f}'(X\mathbf{\beta}_k) - \mathbf{f}'(X\mathbf{\beta^*})||_2^2 + \Psi \label{eq:thetadiff}, 
\end{align}
where 
\begin{align*}
\Psi = &\frac{\tau_k^2\varphi^2_k }{\lambda^2 } || \mathbf{f}'(X\mathbf{\beta}_k)||_2^2 - \frac{2\tau_k \varphi_k}{\lambda^2 }  \langle (\mathbf{f}'(X\mathbf{\beta}_k) - \mathbf{f}'(X\mathbf{\beta^*}) ),  \mathbf{f}'(X\mathbf{\beta}_k) \rangle  \\
\leq &\frac{\tau_k^2\varphi^2_k }{\lambda^2 } || \mathbf{f}'(X\mathbf{\beta}_k)||_2^2 + \frac{2\tau_k \varphi_k}{\lambda^2 }  || (\mathbf{f}'(X\mathbf{\beta}_k) - \mathbf{f}'(X\mathbf{\beta^*}) )||_2  ||\mathbf{f}'(X\mathbf{\beta}_k) ||_2. 
\end{align*}
With $ |\varphi_k|\leq |x_m^T\rho_k|\leq ||x_m||_2||\hat{\theta}_k - \theta^*||_2  =  \frac{ ||x_m||_2}{\lambda}||\mathbf{f}'(X\mathbf{\beta}_k) - \mathbf{f}'(X\mathbf{\beta^*})||_2 $,
\begin{align} \nonumber
\Psi  &\leq \frac{\tau_k^2||x_m||_2^2}{\lambda^4 } || \mathbf{f}'(X\mathbf{\beta}_k)||_2^2 ||\mathbf{f}'(X\mathbf{\beta}_k) - \mathbf{f}'(X\mathbf{\beta^*})||_2^2 + \frac{2\tau_k||x_m||_2 }{\lambda^3 } ||\mathbf{f}'(X\mathbf{\beta}_k) ||_2 || (\mathbf{f}'(X\mathbf{\beta}_k) - \mathbf{f}'(X\mathbf{\beta^*}) )||_2^2  \\
&= \bigg(\frac{\tau_k^2||x_m||_2^2}{\lambda^4 } || \mathbf{f}'(X\mathbf{\beta}_k)||_2^2+ \frac{2\tau_k||x_m||_2 }{\lambda^3 } ||\mathbf{f}'(X\mathbf{\beta}_k) ||_2 \bigg)  || (\mathbf{f}'(X\mathbf{\beta}_k) - \mathbf{f}'(X\mathbf{\beta^*}) )||_2^2 . \label{eq:psiineq}
\end{align}
With~\eqref{eq:thetadiff} and~\eqref{eq:psiineq}, 
\begin{align*}
||\theta_k - \theta^*||^2_2 \leq &   \bigg(\frac{1}{\lambda^2} +\frac{2\tau_k||x_m||_2 }{\lambda^3 } ||\mathbf{f}'(X\mathbf{\beta}_k) ||_2+ \frac{\tau_k^2||x_m||_2^2}{\lambda^4 } || \mathbf{f}'(X\mathbf{\beta}_k)||_2^2 \bigg)  || (\mathbf{f}'(X\mathbf{\beta}_k) - \mathbf{f}'(X\mathbf{\beta^*}) )||_2^2 \\
=&\frac{1}{\lambda^2}\bigg(1 +\frac{2\tau_k||x_m||_2 }{\lambda } ||\mathbf{f}'(X\mathbf{\beta}_k) ||_2+ \frac{\tau_k^2||x_m||_2^2}{\lambda^2 } || \mathbf{f}'(X\mathbf{\beta}_k)||_2^2 \bigg)  || (\mathbf{f}'(X\mathbf{\beta}_k) - \mathbf{f}'(X\mathbf{\beta^*}) )||_2^2 .
\end{align*}

$L$ is the Lipschitz continuous constant of $\mathbf{f}'$. As $\mathbf{f}$ is convex and smooth, there always is a point, e.g. $\bar{\mathbf{\beta}}$, with $\mathbf{f}'(X\bar{\mathbf{\beta}}) = \mathbf{0}$. Hence, 
\begin{align*} ||\mathbf{f}'(X\mathbf{\beta}_k) -\mathbf{f}'(X\bar{\mathbf{\beta}})||_2 = ||\mathbf{f}'(X\mathbf{\beta}_k) -\mathbf{0}||_2=||\mathbf{f}'(X\mathbf{\beta}_k)||_2 \leq L||\mathbf{\beta}_k -\bar{\mathbf{\beta}}||_{\Sigma}.
\end{align*}
With $\mathbf{\beta}_k$ and $\bar{\mathbf{\beta}}$ being finite values, $||\mathbf{f}'(X\mathbf{\beta}_k)||_2$ is also a finite value. We can make the following assumption:
\begin{align*} 
&\bigg(1 +\frac{2\tau_k||x_m||_2 }{\lambda } ||\mathbf{f}'(X\mathbf{\beta}_k) ||_2+ \frac{\tau_k^2||x_m||_2^2}{\lambda^2 } || \mathbf{f}'(X\mathbf{\beta}_k)||_2^2 \bigg) \\
=&\bigg(1 + \frac{\tau_k||x_m||_2}{\lambda } || \mathbf{f}'(X\mathbf{\beta}_k)||_2 \bigg) ^2 
 \leq  M .
\end{align*}
Hence, \[||\theta_k - \theta^*||^2_2 \leq  \frac{M}{\lambda^2}|| (\mathbf{f}'(X\mathbf{\beta}_k) - \mathbf{f}'(X\mathbf{\beta^*}) )||_2^2 \leq \frac{LM}{\lambda^2} || \beta_k - \beta^* ||^2_{\Sigma} .\] (Q.E.D.)

%
%

\section*{Appendix C. Proof of Theorem 4, Lemma 3, and Theorem 5}
\noindent\textbf{Theorem 4} {\it Assuming that the time complexity for one coordinate minimization operation is $O(u)$, the time complexity for dynamic screening  is  $O\Big(u\frac{\bar{L}^2}{\gamma^2}\big( p\log\frac{G_0}{\varepsilon_D} + |\bar{ \mathcal{A}}| \log \frac{\varepsilon_D}{\varepsilon} \big)\Big)$. Here $G_0 = P(\beta_0) - P(\beta^*)$, and $\varepsilon_D$ is the  accuracy of the objective function value for the last feature screening operation. }

\noindent\textbf{Proof}: The computation of dynamic screening has two main phases, feature screening and accuracy pursuing, their time complexity denoted by $T_a$ and $T_b$ respectively. Let $G_d = P(\beta_{d}) - P(\beta^*)$ represent the primal accuracy for the screening of the $d$th feature, and $p_d$  is the size of feature set after $d$ features have been removed with the screening procedure, i.e, $p_d = p - d$. $O(Ku)$ is the  complexity for $K$ CM iterations, and we need  $np_d$ to compute the duality gap for each outer loop. With Lemma 1, $\psi_d =  \frac{p_d\bar{L}_d^2}{p_d\bar{L}_d^2 + \gamma^2}$. Here $\bar{L}_d = \sqrt{\sigma_{d \max}}L$, $\sigma_{d\max}$ is the largest eigenvalue of $X_d^TX_d$, and $X_d$ is the design matrix after the screening of $d$ features. Assume $Z=p - |\bar{ \mathcal{A}}|$, which is the total number of features that can be removed. By Lemma 1, we have

\begin{align*}
T_a \leq  \sum_{d=1}^{Z}  \frac{\log_{\psi_{d-1}} \frac{G_d}{G_{d-1}}}{K}  (Ku + np_{d-1}) ,
\end{align*}
and 
\begin{align*}
T_b \leq  u\log_{\psi_Z} \frac{\varepsilon}{G_Z} .
\end{align*}

Hence, the time complexity  for dynamic screening adds up:
\begin{align}
T &= T_a + T_b \nonumber\\ 
&\leq  \sum_{d=1}^{Z}  \frac{\log_{\psi_{d-1}} \frac{G_d}{G_{d-1}}}{K}  (Ku + np_d) + u\log_{\psi_Z} \frac{\varepsilon}{G_Z}  \nonumber \\
&= u\sum_{d=1}^{Z}  \log_{\psi_{d-1}} \frac{G_d}{G_{d-1}} +  \frac{n}{K}\sum_{d=1}^{Z} p_{d-1}\log_{\psi_{d-1}} \frac{G_d}{G_{d-1}}   +u\log_{\psi_Z} \frac{\varepsilon}{G_Z}   \nonumber \\  
&=  u\sum_{d=1}^{Z}  \log_{\psi_{d-1}} \frac{G_d}{G_{d-1}}+u\log_{\psi_Z} \frac{\varepsilon}{G_Z}+ \frac{n}{K} \sum_{d=1}^{Z}  p_{d-1}\log_{\psi_{d-1}} \frac{G_d}{G_{d-1}} .
\end{align} \label{eq:tabe}

Following the proof of Theorem 3 in~\cite{Li2017},
\begin{align*}
\log_{\psi_{d-1}} \frac{G_d}{G_{d-1}} & \leq ( 1+ \frac{p_{d-1}\bar{L}_{d-1}^2 }{ \gamma^2} )\log\frac{G_{d-1}}{G_d} ,\\
\log_{\psi_Z} \frac{\varepsilon}{G_Z}  &\leq (1+\frac{|\bar{ \mathcal{A}}|\bar{L}_Z^2}{\gamma^2}) \log\frac{G_Z}{\varepsilon} .
\end{align*}
For the first two terms in (29), denoted by $T_1$, we have 
\begin{align*}
T_1&=  u\sum_{d=1}^{Z}  \log_{\psi_{d-1}} \frac{G_d}{G_{d-1}}+u\log_{\psi_Z} \frac{\varepsilon}{G_Z}\\
&\leq  u\sum_{d=1}^{Z} (1+\frac{p_{d-1}\bar{L}_{d-1}^2}{\gamma^2})\log\frac{G_{d-1}}{G_d}+ u(1+\frac{|\bar{ \mathcal{A}}|\bar{L}_Z^2}{\gamma^2}) \log\frac{G_Z}{\varepsilon}\\
&\leq u\log\frac{G_0}{\varepsilon} + \frac{u\bar{L}^2}{\gamma^2}\log \bigg[ \big( \Pi_{d=1}^{Z} \frac{G_{d-1}^{p_{d-1}}}{G_d^{p_{d-1}}} \big) \frac{G_Z^{|\bar{ \mathcal{A}}|}}{\varepsilon^{|\bar{ \mathcal{A}}|}} \bigg] \\
&= u\log\frac{G_0}{\varepsilon} + \frac{u\bar{L}^2}{\gamma^2}\log \frac{G_0^p}{\varepsilon^{|\bar{ \mathcal{A}}| } \Pi_{d=1}^{Z} G_d } \\
&=  u\log\frac{G_0}{\varepsilon} + \frac{u\bar{L}^2}{\gamma^2}\Big((p-|\bar{ \mathcal{A}}|)\log \frac{G_0}{\bar{G}}  + |\bar{ \mathcal{A}}|\log \frac{G_0}{\varepsilon} \Big) .
\end{align*}
Here $\bar{G} = \big(\Pi_{d=1}^{Z} G_d \big)^{\frac{1}{p-|\bar{ \mathcal{A}}|}}$.  For the remaining terms in (29), denoted by $T_2$, we have 
\begin{align*}
T_2 &=
\frac{n}{K}\sum_{d=1}^{Z}  \log_{\psi_{d-1}} \frac{G_d}{G_{d-1}} p_{d-1} \\
& \leq  \frac{n}{K}\sum_{d=1}^{Z}(p_{d-1}+\frac{p_{d-1}^2 \bar{L}_{d-1}^2}{\gamma^2})\log\frac{G_{d-1}}{G_d} \\
&\leq  \frac{n}{K} \Big( \log \frac{G_0^p}{\bar{G}^{p-|\bar{ \mathcal{A}}|} G_Z^{|\bar{ \mathcal{A}}|}}  + \frac{\bar{L}^2}{\gamma^2}\log \frac{G_0^{p^2}}{\tilde{G}^{p^2-|\bar{ \mathcal{A}}|^2} G_Z^{|\bar{ \mathcal{A}}|^2}} \Big) \ ,
\end{align*}
where $\tilde{G} = \big(\Pi_{d=1}^{Z} G_d^{p^2_{d-1} -p^2_{d}}\big)^{\frac{1}{p^2-|\bar{ \mathcal{A}}|^2}}$ .

The primal accuracy for feature screening is increasing, i.e. $\forall d \leq Z, \ G_{d} \geq G_{Z}$, and this leads to
\begin{align*}
\bar{G} = \big(\Pi_{d=1}^{Z} G_d \big)^{\frac{1}{p-|\bar{ \mathcal{A}}|}} \geq \big(\Pi_{d=1}^{Z} G_Z \big)^{\frac{1}{p-|\bar{ \mathcal{A}}|}} = G_Z ,
\end{align*}
and similarly,
\begin{align*}
\tilde{G} = \big(\Pi_{d=1}^{Z} G_d^{p^2_{d-1} -p^2_{d}}\big)^{\frac{1}{p^2-|\bar{ \mathcal{A}}|^2}} \geq \big(\Pi_{d=1}^{Z} G_Z^{p^2_{d-1} -p^2_{d}}\big)^{\frac{1}{p^2-|\bar{ \mathcal{A}}|^2}} = G_Z .
\end{align*}
\begin{align*}
T \leq & T_1 + T_2 \\
\leq & u\log\frac{G_0}{\varepsilon} + \frac{u\bar{L}^2}{\gamma^2}\Big((p-|\bar{ \mathcal{A}}|)\log \frac{G_0}{\bar{G}}  + |\bar{ \mathcal{A}}|\log \frac{G_0}{\varepsilon} \Big) + \frac{n}{K} \Big( \log \frac{G_0^p}{G_Z^{p-|\bar{ \mathcal{A}}|} G_Z^{|\bar{ \mathcal{A}}|}}  + \\ &\frac{\bar{L}^2}{\gamma^2}\log \frac{G_0^{p^2}}{G_Z^{p^2-|\bar{ \mathcal{A}}|^2} G_Z^{|\bar{ \mathcal{A}}|^2}} \Big) \\
\leq &  u\log\frac{G_0}{\varepsilon} + \frac{u\bar{L}^2}{\gamma^2}\Big((p-|\bar{ \mathcal{A}}|)\log \frac{G_0}{G_Z}  + |\bar{ \mathcal{A}}|\log \frac{G_0}{\varepsilon} \Big) + \frac{n}{K} \Big( p\log \frac{G_0}{G_Z}  + \frac{p^2\bar{L}^2}{\gamma^2}\log \frac{G_0}{G_Z} \Big) .
\end{align*}

We can set $K = Cp$, where $C$ is a constant. We have
\begin{align*}
T &\leq  u\log\frac{G_0}{\varepsilon} + \frac{u\bar{L}^2}{\gamma^2}\Big((p-|\bar{ \mathcal{A}}|)\log \frac{G_0}{G_Z}  + |\bar{ \mathcal{A}}|\log \frac{G_0}{\varepsilon} \Big) + \frac{n}{K} \Big( p\log \frac{G_0}{G_Z}  + \frac{p^2\bar{L}^2}{\gamma^2}\log \frac{G_0}{G_Z} \Big)\\
& = u\log\frac{G_0}{\varepsilon} + \frac{u\bar{L}^2}{\gamma^2}\Big((p-|\bar{ \mathcal{A}}|)\log \frac{G_0}{G_Z}  + |\bar{ \mathcal{A}}|\log \frac{G_0}{\varepsilon} \Big) + \frac{n}{C} \Big( \log \frac{G_0}{G_Z}  + \frac{p\bar{L}^2}{\gamma^2}\log \frac{G_0}{G_Z} \Big) \\
&= \Big(u\frac{\bar{L}^2}{\gamma^2}(p-|\bar{ \mathcal{A}}|+\frac{p}{C}) + \frac{n}{C} + \frac{n}{C}\frac{p\bar{L}^2}{\gamma^2} \Big)\log\frac{G_0}{G_Z} + u(1+\frac{\bar{L}^2}{\gamma^2}|\bar{ \mathcal{A}}|)\log \frac{G_0}{\varepsilon} \\
&= \Big(u\frac{\bar{L}^2}{\gamma^2}(p+\frac{p}{C}) + u+ \frac{n}{C} + \frac{n}{C}\frac{p\bar{L}^2}{\gamma^2} \Big)\log\frac{G_0}{G_Z} + u(1+\frac{\bar{L}^2}{\gamma^2}|\bar{ \mathcal{A}}|)\log \frac{G_Z}{\varepsilon} \\
&= up\eta\frac{\bar{L}^2}{\gamma^2} \log\frac{G_0}{G_Z} + u\frac{\bar{L}^2}{\gamma^2}|\bar{ \mathcal{A}}|\log \frac{G_Z}{\varepsilon} +  u\log \frac{G_Z}{\varepsilon}+ (u + \frac{n}{C} ) \log\frac{G_0}{G_Z} .
\end{align*}

Here $\eta = 1+\frac{1}{C} + \frac{u}{Cn}$. With $\epsilon_D = G_Z$, ignoring the last two items, the complexity of dynamic screening can be simplified  as $O\Big( u\frac{\bar{L}^2}{\gamma^2}\big( p\log\frac{G_0}{\varepsilon_D} + |\bar{ \mathcal{A}}| \log \frac{\varepsilon_D}{\varepsilon} \big) \Big)$. (Q.E.D.)

\noindent\textbf{Lemma 3} {\it With $O(u)$ as the complexity for the base operation of coordinate minimization of the LASSO problem with a $\gamma$-convex loss function, $p_A$ the total number of features involved in ADD operations, and $p_{p_A}$ the size of the active set when IsADD is set to false, the complexity for the feature recruiting phase in SAIF is 
	\begin{align} \nonumber
	\frac{Ku+pn}{K} \Big( \Upsilon + \frac{\bar{L}^2}{\gamma^2} \Phi+ p_{p_A} \frac{\bar{L}^2}{\gamma^2} \log \frac{\bar{Q}}{Q_{p_A}(\beta_{p_A})} \Big),
	\end{align} 
	where 
	\begin{align*} 
	&\bar{Q} = \big(\Pi_{d=1}^{p_A-1}Q_{d+1}(\beta_d)^{p_{d+1}-p_d}\big)^{\frac{1}{p_{A}}} , \ \  \Upsilon = \log\big( \Pi_{d=1}^{p_A -1 } \frac{Q_{d+1}(\beta_d)}{Q_d(\beta_d)^{\frac{p_d}{p_{d+1}}}} \frac{1}{Q_{p_A}(\beta_{p_A})} \big) , \\
	&\Phi =  \log\big( \Pi_{d=1}^{p_A -1} \frac{Q_{d+1}(\beta_d)^{p_{d}}}{Q_d(\beta_d)^{p_d}}\big). 
	\end{align*}
}
\noindent\textbf{Proof}:
Let $p_A$ be the total number of features involved in the ADD operations. $Q_d(\beta) = P_d(\beta) - P_d(\beta^*_d)$. We add up the time complexity of the outer loops regarding each added feature. O($Ku$) is the time complexity for $K$ base CM operations; $O(np)$ is the computation complexity for duality gap and the ADD operation in one iteration of outer loop. We have
\begin{align*}
	T_a \leq & \sum_{d=1}^{p_A}  \frac{\log_{\psi_{d}} \frac{Q_d(\beta_d)}{Q_{d}(\beta_{d-1})}}{K} ( Ku + np) \\
	=& \frac{Ku+np}{K}  \sum_{d=1}^{p_A} \log_{\psi_{d}} \frac{Q_d(\beta_d)}{Q_{d}(\beta_{d-1})} \\
	=& \frac{Ku+np}{K} \Big( -\log_{\psi_1}Q_1(\beta_0) + \sum_{d=1}^{p_A-1}\big( \log_{\psi_d}Q_d(\beta_d) - \log_{\psi_{d+1}} Q_{d+1}(\beta_d)\big) +\log_{\psi_{p_A}} Q_{p_A}(\beta_{p_A}) \Big)\\
	=&  \frac{Ku+np}{K} \Big( -\log_{\psi_1}Q_1(\beta_0) + \sum_{d=1}^{p_A-1}\big( \log_{\psi_d}Q_d(\beta_d) - \log_{\psi_{d+1}} Q_{d+1}(\beta_d)\big) +\log_{\psi_{p_A}} Q_{p_A}(\beta_{p_A}) \Big)\\
	= & \frac{Ku+np}{K} \Big( \log_{\psi_{p_A}} Q_{p_A}(\beta_{p_A})-\log_{\psi_1}Q_1(\beta_0) + \sum_{d=1}^{p_A-1}\log_{\psi_{d+1}} \frac{Q_d(\beta_d)^{\frac{\log\psi_{d+1}}{\log \psi_d}}}{Q_{d+1}(\beta_d)} \Big) \ .
\end{align*}
With 	$\psi_{d+1} = \frac{ \gamma^2 }{ p_{d+1}\bar{L}_{d+1}^2 + \gamma^2}$, by following the proof of Theorem 3 in~\cite{Li2017},
\begin{align*}
\log_{\psi_{d+1}} \frac{Q_t(\beta_d)^{\frac{\log\psi_{d+1}}{\log \psi_d}}}{Q_{d+1}(\beta_d)}&\leq \frac{p_{d+1}\bar{L}_{d+1}^2 + \gamma^2}{ \gamma^2} \log \frac{Q_{d+1}(\beta_d)}{ Q_t(\beta_d)^{\frac{\log\psi_{d+1}}{\log \psi_d}}} .
\end{align*}
As $\forall d,  \bar{L}_{d+1} \leq \bar{L}$, the time complexity for the feature recruiting phase in SAIF 
	
\begin{align*}
		T_a \leq & \frac{Ku+np}{K} \Big( -(1+p_{p_A}\frac{\bar{L}^2}{\gamma^2})\log Q_{p_A}(\beta_{p_A})-\log_{\psi_1}Q_1(\beta_0) + \\ &\sum_{d=1}^{p_A-1}(1+p_{d+1}\frac{\bar{L}^2}{\gamma^2})\log \frac{Q_{d+1}(\beta_d)}{Q_d(\beta_d)^{\frac{\log\psi_{d+1}}{\log \psi_d}}} \Big) \ .
\end{align*}
With  $\bar{L}_{d+1} \approx \bar{L}_{d}$, $\frac{\log\psi_{d+1}}{\log \psi_d} = \frac{\log\frac{ \gamma^2 }{ p_{d+1}\bar{L}_{d+1}^2 + \gamma^2}}{\log \frac{ \gamma^2 }{ p_{d}\bar{L}_{d}^2 + \gamma^2}} =  \frac{\log(1 - \frac{ p_{d+1}\bar{L}_{d+1}^2 }{ p_{d+1}\bar{L}_{d+1}^2 + \gamma^2})}{\log(1- \frac{  p_{d}\bar{L}_{d}^2 }{ p_{d}\bar{L}_{d}^2 + \gamma^2})} \approx   \frac{p_{d}}{ p_{d+1}}$, we have	
	
\begin{align*}
	T_a \leq &   \frac{Ku+np}{K} \Big( -(1+p_{p_A}\frac{\bar{L}^2}{\gamma^2})\log Q_{p_A}(\beta_{p_A})-\log_{\psi_1}Q_1(\beta_0) + \sum_{d=1}^{p_A-1}(1+p_{d+1}\frac{\bar{L}^2}{\gamma^2})\log \frac{Q_{d+1}(\beta_d)}{Q_d(\beta_d)^{\frac{p_d}{p_{d+1}}}} \Big)\\
	\leq & \frac{Ku+np}{K} \Big( \log\big( \Pi_{d=1}^{p_A -1 } \frac{Q_{d+1}(\beta_d)}{Q_d(\beta_d)^{\frac{p_d}{p_{d+1}}}} \frac{1}{Q_{p_A}(\beta_{p_A})} \big)  + \frac{\bar{L}^2}{\gamma^2}  \log\big( \Pi_{d=1}^{p_A -1} \frac{Q_{d+1}(\beta_d)^{p_{d+1}}}{Q_d(\beta_d)^{p_d}}\big) \frac{1}{Q_{p_A}(\beta_{p_A})^{p_{p_A}}} \Big)\\
	= & \frac{Ku+np}{K} \Big( \log\big( \Pi_{d=1}^{p_A -1 } \frac{Q_{d+1}(\beta_d)}{Q_d(\beta_d)^{\frac{p_d}{p_{d+1}}}} \frac{1}{Q_{p_A}(\beta_{p_A})} \big)  + \frac{\bar{L}^2}{\gamma^2}  \log\big( \Pi_{d=1}^{p_A -1} \frac{Q_{d+1}(\beta_d)^{p_{d}}}{Q_d(\beta_d)^{p_d}}\big)+   \\
	& \frac{\bar{L}^2}{\gamma^2} \log \frac{\Pi_{d=1}^{p_A-1}Q_{d+1}(\beta_d)^{p_{d+1}-p_d}}{Q_{p_A}(\beta_{p_A})^{p_{p_A}}} \Big) \\ 
	= & \frac{Ku+np}{K} \Big( \log\big( \Pi_{d=1}^{p_A -1 } \frac{Q_{d+1}(\beta_d)}{Q_d(\beta_d)^{\frac{p_d}{p_{d+1}}}} \frac{1}{Q_{p_A}(\beta_{p_A})} \big)  + \frac{\bar{L}^2}{\gamma^2}  \log\big( \Pi_{d=1}^{p_A -1} \frac{Q_{d+1}(\beta_d)^{p_{d}}}{Q_d(\beta_d)^{p_d}}\big)+ \\ & p_{p_A} \frac{\bar{L}^2}{\gamma^2} \log \frac{\bar{Q}}{Q_{p_A}(\beta_{p_A})} \Big) .
\end{align*}

Here 
\begin{align*}
	\bar{Q} = \big(\Pi_{d=1}^{p_A-1}Q_{d+1}(\beta_d)^{p_{d+1}-p_d}\big)^{\frac{1}{p_{p_A}}} .
\end{align*}

Let 
\begin{align*}
	\Upsilon = \log\big( \Pi_{d=1}^{p_A -1 } \frac{Q_{d+1}(\beta_d)}{Q_d(\beta_d)^{\frac{p_d}{p_{d+1}}}} \frac{1}{Q_{p_A}(\beta_{p_A})} \big) ,
\end{align*}
and 
\begin{align*}
	\Phi =  \log\big( \Pi_{d=1}^{p_A -1} \frac{Q_{d+1}(\beta_d)^{p_{d}}}{Q_d(\beta_d)^{p_d}}\big).
\end{align*}
This results in
\begin{align*}
	T_a \leq  \frac{Ku+np}{K} \Big( \Upsilon + \frac{\bar{L}^2}{\gamma^2} \Phi+ p_{p_A} \frac{\bar{L}^2}{\gamma^2} \log \frac{\bar{Q}}{Q_{p_A}(\beta_{p_A})} \Big) . \ \text{(Q.E.D.)}
\end{align*}

\noindent\textbf{Theorem 5} {\it With $O(u)$ as the complexity for the base operation of coordinate minimization of the LASSO problem  with a $\gamma$-convex loss function, 
	the time complexity for SAIF is $O\bigg(u\frac{\bar{L}^2}{\gamma^2} \big( \bar{p} \log \frac{\bar{Q}}{\varepsilon_D} + \bar{p} p_A + |\bar{ \mathcal{A}}|\log \frac{\varepsilon_D}{\varepsilon} \big) \bigg) $. Here $p_A$ is the total number of features involved in ADD operations, $\bar{p}$ is the maximum size of the active set during the algorithm iterations, $\bar{Q}$ is the geometric mean of the accuracies of the sub-problem objective function values corresponding to each ADD operation, and $\varepsilon_D$ is the accuracy of the objective function value for the last feature DEL operation. } 

\noindent\textbf{Proof}:
 $T_b$ denotes the time consumed by both inactive feature screening  and accuracy pursuing phases. The inactive feature screening and accuracy pursuing phases are similar to dynamic screening. We simplify the derivations by following the steps and techniques used in the analysis for dynamic screening. Let $Z_D$ be the total number of features removed with screening after the stop of ADD operations. We have 
  
\begin{align*}
T_b &=\sum_{d=1}^{Z_D}  \frac{\log_{\psi_{d-1}} \frac{G_d}{G_{d-1}}}{K} ( Ku + np_{d-1}) + u\log_{\psi_{Z_D}} \frac{\varepsilon}{G_{Z_D}}\\
&=u\sum_{d=1}^{Z_D} \log_{\psi_{d-1}} \frac{G_d}{G_{d-1}} + u\log_{\psi_{Z_D}} \frac{\varepsilon}{G_{Z_D}} + \frac{n}{K}\sum_{d=1}^{Z_D} p_{d-1}\log_{\psi_{d-1}} \frac{G_d}{G_{d-1}} .
\end{align*}

The first two terms can be written as 
\begin{align*}
T_{b1}&= u\sum_{d=1}^{Z_D} \log_{\psi_{d-1}} \frac{G_d}{G_{d-1}} + u\log_{\psi_{Z_D}} \frac{\varepsilon}{G_{Z_D}} \\
&\leq u\log\frac{G_{p_{p_A}}}{\varepsilon} + \frac{u\bar{L}^2}{\gamma^2}\Big((p_{p_A}-|\bar{ \mathcal{A}}|)\log \frac{G_{p_{p_A}}}{\bar{G}}  + |\bar{ \mathcal{A}}|\log \frac{G_{p_{p_A}}}{\varepsilon} \Big) .
\end{align*}
Here $\bar{G} = \big(\Pi_{d=1}^{Z_D} G_d \big)^{\frac{1}{p_{p_A}-|\bar{ \mathcal{A}}|}}$.

\begin{align*}
T_{b2}&= \frac{n}{K}\sum_{d=1}^{Z_D} p_{d-1}\log_{\psi_{d-1}} \frac{G_d}{G_{d-1}} \\
&\leq  \frac{n}{K}\sum_{d=1}^{Z_D}(p_{d-1}+\frac{p_{d-1}^2 \bar{L}^2}{\gamma^2})\log\frac{G_{d-1}}{G_d} \\
&=  \frac{n}{K} \Big( \log \frac{G_{p_{p_A}}^{p_{p_A}}}{\bar{G}^{p_{p_A}-|\bar{ \mathcal{A}}|} G_{Z_D}^{|\bar{ \mathcal{A}}|}}  + \frac{\bar{L}^2}{\gamma^2}\log \frac{G_{p_{p_A}}^{p_{p_A}^2}}{\tilde{G}^{p_{p_A}^2-|\bar{ \mathcal{A}}|^2} G_{Z_D}^{|\bar{ \mathcal{A}}|^2}} \Big) ,
\end{align*}

where $\tilde{G} = \big(\Pi_{d=1}^{Z_D} G_d^{p^2_{d-1} -p^2_{d}}\big)^{\frac{1}{p_{p_A}^2-|\bar{ \mathcal{A}}|^2}}$ .

Similar to dynamic screening, 
\begin{align*}
 \bar{G} \geq \big(\Pi_{d=1}^{Z_D} G_{Z_D} \big)^{\frac{1}{p_{p_A}-|\bar{ \mathcal{A}}|}}=G_{Z_D} ,
\end{align*}
and 

\begin{align*}
\tilde{G} \geq \big(\Pi_{d=1}^{Z_D} G_{Z_D}^{p^2_{d-1} -p^2_{d}}\big)^{\frac{1}{p_{p_A}^2-|\bar{ \mathcal{A}}|^2}} = G_{Z_D}.
\end{align*}

Thus 
\begin{align*}
T_{b1}& \leq u\log\frac{G_{p_{p_A}}}{\varepsilon} + \frac{u\bar{L}^2}{\gamma^2}\Big((p_{p_A}-|\bar{ \mathcal{A}}|)\log \frac{G_{p_{p_A}}}{G_{Z_D}}  + |\bar{ \mathcal{A}}|\log \frac{G_{p_{p_A}}}{\varepsilon} \Big) ,
\end{align*}
and
\begin{align*}
T_{b2}& \leq \frac{n}{K} \Big( p_{p_A}\log \frac{G_{p_{p_A}}}{ G_{Z_D}}  + \frac{\bar{L}^2}{\gamma^2} p_{p_A}^2\log \frac{G_{p_{p_A}}}{ G_{Z_D}}\Big) .
\end{align*}

We set $K$ proportional to feature size for both feature recruiting and inactive feature screening phases, i.e., $K_I = C p$ and  $K_D = C p_{p_A}$. 
With $G_{p_{p_A}} = Q_{p_A}(\beta_{p_A})$, the time complexity for SAIF can be written as

\begin{align*}
T=& T_a + T_b \\
=& T_a + T_{b1} + T_{b2} \\
\leq &  \frac{K_Iu+np}{K_I} \Big( \Upsilon + \frac{\bar{L}^2}{\gamma^2} \Phi+ p_{p_A} \frac{\bar{L}^2}{\gamma^2} \log \frac{\bar{Q}}{Q_{p_A}(\beta_{p_A})} \Big)+  u\log\frac{G_{p_{p_A}}}{\varepsilon} + \frac{u\bar{L}^2}{\gamma^2}\Big((p_{p_A}-|\bar{ \mathcal{A}}|)\log \frac{G_{p_{p_A}}}{G_{Z_D}} \\& + |\bar{ \mathcal{A}}|\log \frac{G_{p_{p_A}}}{\varepsilon} \Big)  + \frac{n}{K_D} \Big( p_{p_A}\log \frac{G_{p_{p_A}}}{ G_{Z_D}}  + \frac{\bar{L}^2}{\gamma^2} p_{p_A}^2\log \frac{G_{p_{p_A}}}{ G_{Z_D}}\Big)\\
= &(u+\frac{n}{C}) \Big(\Upsilon + \frac{\bar{L}^2}{\gamma^2} \Phi+ p_{p_A} \frac{\bar{L}^2}{\gamma^2} \log \frac{\bar{Q}}{G_{p_{p_A}}}\Big) + u\log\frac{G_{p_{p_A}}}{\varepsilon} + \frac{u\bar{L}^2}{\gamma^2}\Big( p_{p_A}\log\frac{G_{p_{p_A}}}{G_{Z_D}} + |\bar{ \mathcal{A}}|\log \frac{G_{Z_D}}{\varepsilon} \Big)\\
&+ \frac{n}{C} \Big( \log \frac{G_{p_{p_A}}}{ G_{Z_D}}  + \frac{\bar{L}^2}{\gamma^2} p_{p_A}\log \frac{G_{p_{p_A}}}{ G_{Z_D}}\Big) \\
= & p_{p_A}(u+\frac{n}{C})\frac{\bar{L}^2}{\gamma^2} \log \frac{\bar{Q}}{G_{Z_D}}+ u\frac{\bar{L}^2}{\gamma^2}|\bar{ \mathcal{A}}|\log \frac{G_{Z_D}}{\varepsilon} +(u+\frac{n}{C}) \frac{\bar{L}^2}{\gamma^2} \Phi+ (u+\frac{n}{C})\Upsilon+u\log\frac{G_{p_{p_A}}}{\varepsilon}+ \\ &\frac{n}{C} \log \frac{G_{p_{p_A}}}{ G_{Z_D}} .
\end{align*}

Let $\eta = 1+\frac{n}{uC}$, $\bar{p} =\max_{d: 1\leq d \leq p_A} p_d$, and $\mu = \max_{d: 1\leq d \leq p_A-1} \eta\log\frac{Q_{d+1}(\beta_d)}{Q_d(\beta_d)}$,  then we have
\begin{align*}
\eta\Phi& =  \eta\log\big( \Pi_{d=1}^{p_A -1} \frac{Q_{d+1}(\beta_d)^{p_{d}}}{Q_d(\beta_d)^{p_d}}\big)
=\sum_{d=1}^{p_A-1}p_d\eta\log\frac{Q_{d+1}(\beta_d)}{Q_d(\beta_d)}\\
&\leq \mu \sum_{d=1}^{p_A-1}p_d  \leq \mu \bar{p} p_A,
\end{align*}

and 

\begin{align*}
T \leq  u \eta p_{p_A}\frac{\bar{L}^2}{\gamma^2} \log \frac{\bar{Q}}{G_{Z_D}}+ u\frac{\bar{L}^2}{\gamma^2}|\bar{ \mathcal{A}}|\log \frac{G_{Z_D}}{\varepsilon} + u \mu \bar{p} p_A+  u \eta\Upsilon+u\log\frac{G_{p_{p_A}}}{\varepsilon}+  &\frac{n}{C} \log \frac{G_{p_{p_A}}}{ G_{Z_D}} .
\end{align*}

Let $\epsilon_D =  G_{Z_D}$,   the time complexity for SAIF can be simplified as  $O\bigg(u\frac{\bar{L}^2}{\gamma^2} \big(\bar{p} \log \frac{\bar{Q}}{\varepsilon_D} +\bar{p} p_A + |\bar{ \mathcal{A}}|\log \frac{\varepsilon_D}{\varepsilon} \big) \bigg) $. (Q.E.D.)


\section*{Appendix D. Proofs of Theorems 6 and 7}

\noindent\textbf{Theorem 6} {\it If $D$ can be transformed into a diagonal matrix with a column transformation matrix T, i.e. $\tilde{D} = D T$, and $\tilde{D}$ is a diagonal matrix, then
	
	a) the problem~\eqref{eq:genls} is equivalent to    
	\begin{align*}
	& \tilde{P}:\  \  \min_{\tilde{\beta}, b}\sum_{j=1}^n f\Big(\sum_{i=1}^{p-1} \tilde{x}_{ji} \tilde{\beta}_i + \tilde{x}_{jp} b, y_j\Big) + \lambda ||\tilde{\beta}||_1, 
	\end{align*}
	
	where $\tilde{X} = XT$, and the solution relationship is $\beta^* = T \bigl[\begin{smallmatrix}
	\tilde{\beta}^* \\b^*
	\end{smallmatrix} \bigr]$; 
	
	b) a dual form of~\eqref{eq:fusedtp} is 
	\begin{align*}
	&\tilde{D}: \ \  \min_{\bar{\theta} \in \Omega} -\sum_{j=1}^{n} f^*(-\lambda \bar{\theta}_j), \,\, \Omega = \big\{\bar{\theta}:  |\bar{x}_i^T\bar{\theta}|  \leq 1, \forall i \in \{1,...,p-1\} \big\}. 
	\end{align*}
	
	Here $\bar{X} = \tilde{X}_{-p}$, and $H =\left[
	\begin{array}{c}
	I \\
	h 
	\end{array}
	\right]  $, $ h = \big[-\frac{\bar{x}_{1,p}}{\bar{x}_{n,p}}, ...,-\frac{\bar{x}_{n-1,p}}{\bar{x}_{n,p}}\big]$.   $\bar{\theta} = H\theta_{-p} $ , and $\theta = -\frac{\mathbf{f}'\big(\tilde{X}\bigl[\begin{smallmatrix}
		\tilde{\beta}^* \\b^* \end{smallmatrix}\bigr]\big)}{\lambda}$;

	c) $\lambda_{max} = \max_{i \in \{1,...,p-1\}} \big|\bar{x}_i^T\mathbf{f}'(\tilde{X}\bigl[\begin{smallmatrix}
	\mathbf{0} \\b \end{smallmatrix}\bigr])\big|$, and $\bar{x}_i$ is the $i$th column of $\bar{X}$. 
}

\noindent\textbf{Proof}:
a) The dual form for fused LASSO~\citep{Ren2017} is 
\begin{align*}
D_1: \  \  & \sup_{\theta} -\sum_{j=1}^n f^*(-\lambda \theta_j) \\
& s.t. \quad X^T\theta = D^T u \\
& \ \quad \quad ||u||_{\infty} \leq 1.
\end{align*}
Here the primal and dual relation for the corresponding  optima is $\theta^* = -\frac{\mathbf{f}(X\beta^*)}{\lambda}$.

With the transformation matrix $T$,  $\tilde{X} = XT$; and  $\bar{D} = DT$ is a diagonal matrix, with the corresponding elements being either 1 or 0 and the last column being an all-zero column. The dual form becomes
\begin{align}
D_2: \  \  & \sup_{\theta} -\sum_{j=1}^n f^*(-\lambda \theta_j) \\ 
& s.t. \quad |\tilde{x}_{i}^T \theta| \leq 1, \forall i, 1 \leq i \leq p-1 \\
& \ \ \quad \quad \tilde{x}_{p}^T \theta =0  \ . \label{eq:dualeq}
\end{align}

We can see that the corresponding transformed primal problem for $D_2$ can be written as: 

\begin{align*}
\tilde{P}: & \  \  \min_{\tilde{\beta}, b}\sum_{j=1} ^n f\Big(\sum_{i=1}^{p-1} \tilde{x}_{ji} \tilde{\beta}_i + \tilde{x}_{jp} b, y_j\Big) + \lambda ||\tilde{\beta}||_1, 
\end{align*}
with $\tilde{X} = XT$. The relationship between the optimal solution to this transformed problem and that to the original problem~\eqref{eq:genls} is $\beta^* = T \bigl[\begin{smallmatrix}
\tilde{\beta}^* \\b^*
\end{smallmatrix} \bigr]$.

b)
With~\eqref{eq:dualeq}, we have $\theta_n = -\frac{\bar{x}_{1,p}}{\bar{x}_{n,p}}\theta_1 - ...-\frac{\bar{x}_{n-1,p}}{\bar{x}_{n,p}}\theta_{n-1} $.
Let $\bar{X} =\tilde{X}_{-p}$, $H =\left[
\begin{array}{c}
I \\
h 
\end{array}
\right]  $, $ h = \big[-\frac{\bar{x}_{1,p}}{\bar{x}_{n,p}}, ...,-\frac{\bar{x}_{n-1,p}}{\bar{x}_{n,p}}\big]$, and $\bar{\theta}=H \theta_{-p}$,  the dual form becomes
\begin{align}
& \ \  \min_{\bar{\theta} \in \Omega_{\lambda}} -\sum_{j=1}^{n} f^*(-\lambda \bar{\theta}_j), \,\, \Omega = \big\{\bar{\theta}:  |\bar{x}_i^T\bar{\theta}|  \leq 1, \forall i \in \{1,...,p-1\} \big\}. \label{eq:gldual-res}
\end{align}
As we have $\theta^* = -\frac{\mathbf{f}(X\beta^*)}{\lambda}$, and $\beta^* = T \bigl[\begin{smallmatrix} \tilde{\beta}^* \\b^* \end{smallmatrix} \bigr]$, therefore $\bar{\theta}^* = -\frac{\big[\mathbf{f}'\big(\tilde{X}\bigl[\begin{smallmatrix}
	\tilde{\beta}^* \\b^* \end{smallmatrix}\bigr]\big)\big]_{-p}}{\lambda}$.

c) As $\lambda_{max}$ is the minimum value of $\lambda$ that  $\tilde{\beta}_1^* = \tilde{\beta}_2^* = ... = \tilde{\beta}_{p-1}^* = 0$, we also have $\max_{i \in \{1,...,p-1\}}|\bar{x}_i^T \bar{\theta} |=1$,  $\Big|\bar{x}_i^T\frac{\big[\mathbf{f}'\big(\tilde{X}\bigl[\begin{smallmatrix}
	\tilde{\beta}^* \\b^* \end{smallmatrix}\bigr]\big)\big]_{-p}}{\lambda_{max}}\Big|=1$. Hence, $\lambda_{max} = \max_{ i\in \{ 1,..., p-1\} } \big|\bar{x}_i^T\mathbf{f}'\big(\tilde{X}\bigl[\begin{smallmatrix}
\mathbf{0} \\b \end{smallmatrix}\bigr]\big)\big| $. (Q.E.D.)

\noindent\textbf{Theorem 7} \ {\it For linear regression problems with fused LASSO regularization,  the scaled feasible $\hat{\bar{\theta}}$ for any $\theta$ that is the closest to $\bar{\theta}^*$ is $\hat{\bar{\theta}} = \tau \bar{\theta}$, where
	$\tau=min\big\{max\{\frac{\mathbf{y}^T\bar{\theta}}{\lambda ||\bar{\theta}||_2^2}, -\frac{1}{||\bar{X}^T\bar{\theta}||_{\infty}}\}, \\ \frac{1}{||\bar{X}^T\bar{\theta}||_{\infty}}\big\}$. }

\noindent\textbf{Proof}: According to Theorem 6, the dual variables corresponding to  the primal variables $\bigl[\begin{smallmatrix} \tilde{\beta} \\b \end{smallmatrix} \bigr]$ can be denoted by $\bar{\theta} = \{\theta_1,...,\theta_{p-1}\} $ , where $\theta = -\frac{\mathbf{f}'\big(\bar{X}\bigl[\begin{smallmatrix}
	\tilde{\beta} \\b \end{smallmatrix}\bigr]\big)}{\lambda} $. While $\bar{\theta}$ may not be feasible to the dual problem of linear regression. With a projection scalar $\tau$, we try to force $\tau \bar{ \theta}$ closer to $\bar{ \theta}^*$ in the feasible space:
\begin{align}
\tau = &\arg\min_{\tau \in R} \bigg\{\frac{1}{2}||\lambda\tau \bar{\theta} - \mathbf{y}||_2^2 - \frac{1}{2}||\mathbf{y}||_2^2, \  s.t. \  \ |\bar{x}_i^T\tau\bar{\theta}|  \leq 1, \forall i \in \{1,...,p-1\} \bigg\} . \label{eq:taumin}
\end{align}
From the objective function in~\eqref{eq:taumin}, we can derive $\tau = \frac{\mathbf{y}^T\bar{\theta}}{\lambda ||\bar{\theta}||_2^2} $ to reach the minimum distance to $\bar{\theta}^*$ if no constraint is imposed on $\tau$. Therefore we need to estimate the range of $\tau$ to determine the minimum. From the constrained region $\big\{ |\bar{x}_i^T\tau\bar{\theta}|  \leq 1, \forall i \in \{1,...,p-1\} \big\}$, the range for $\tau$ is $\big[-\frac{1}{||\bar{X}^T\bar{\theta}||_{\infty}}, \frac{1}{||\bar{X}^T\bar{\theta}||_{\infty}} \big]$. Thus $\tau=min\big\{max\{\frac{\mathbf{y}^T\bar{\theta}}{\lambda ||\bar{\theta}||_2^2}, -\frac{1}{||\bar{X}^T\bar{\theta}||_{\infty}}\}, \frac{1}{||\bar{X}^T\bar{\theta}||_{\infty}}\big\}$. (Q.E.D.)
\bibliography{ref}

\end{document}